\documentclass{style/naturep}

\usepackage[T1]{fontenc}
\usepackage[utf8]{inputenc}  

\usepackage{mathptmx}

\usepackage{amssymb}
\usepackage{amsmath}
\usepackage{graphicx}
\usepackage{algorithmicx}
\usepackage{algorithm}
\usepackage{adjustbox}
\usepackage{graphicx}

\usepackage[noend]{algpseudocode}
\usepackage{bbm}
\usepackage{lineno}
\usepackage[margin=0.6in]{geometry}
\usepackage{ragged2e}
\usepackage{longtable}
\usepackage{xurl}     % Allows URL breaks at arbitrary points
\usepackage{hyperref} % Must be loaded after xurl
\hypersetup{breaklinks=true}
\usepackage{anyfontsize}
\usepackage{setspace}
\usepackage{float}
\usepackage{booktabs}
\usepackage{multirow}
\usepackage{blindtext}
\usepackage{tabularx}
\usepackage{caption}
\usepackage{subcaption}
\usepackage{enumitem, kantlipsum}
\usepackage{xspace}
\usepackage{pifont}% http://ctan.org/pkg/pifont

\usepackage{graphicx}
\usepackage{amsmath}
\usepackage{amssymb}
\usepackage{booktabs}
\usepackage{float}
\usepackage{placeins}
\usepackage[table,x11names]{xcolor}

\usepackage{makecell}
\usepackage{bm}
\usepackage{chngcntr}

\definecolor{maroon}{cmyk}{0,0.87,0.68,0.32}
\definecolor{gray}{rgb}{0.3,0.3,0.3}

\newcommand\Heading[1]{
  \noindent\textbf{\Large{#1}}
}

\newcommand\heading[1]{
  \noindent\textbf{\large{#1}}
}

\newcommand\hheading[1]{
  \noindent\textbf{#1}
}

\usepackage[capitalize,noabbrev]{cleveref}
\usepackage{appendix}
\usepackage{cleveref}

\crefname{figure}{Fig.}{Figs.}
\crefname{suppfigure}{Supplementary Fig.}{Supplementary Figs.}

\crefname{section}{Section}{Sections}
\crefname{theorem}{Theorem}{Theorems}
\crefname{lemma}{Lemma}{Lemmas}
\crefname{equation}{Equation}{Equations}
\crefname{proposition}{Proposition}{Propositions}
\crefname{claim}{Claim}{Claims}
\crefname{appendix}{Appendix}{Appendices}
\crefname{algorithm}{Algorithm}{Algorithms}
\crefname{figure}{Figure}{Figs}
\crefname{table}{Table}{Tables}
\crefname{remark}{Remark}{Remarks}
\crefname{definition}{Definition}{Definitions}
\crefname{equation}{Equation}{Equations}
\crefname{corollary}{Corollary}{Corollaries}
\crefname{section}{Method}{Methods}

\DeclareMathAlphabet{\mathcal}{OMS}{cmsy}{m}{n}

\usepackage{cite}

\title{\begin{flushleft}{\begin{spacing}{1}
    A Generative Foundation Model for Chest Radiography
\end{spacing}}\end{flushleft}}

\makeatletter
\let\saved@includegraphics\includegraphics
\AtBeginDocument{\let\includegraphics\saved@includegraphics}

\makeatother

\begin{document}

%%% Cover Page and abstract
\maketitle
\vspace{-20mm}
\begin{spacing}{1.4}
\noindent Yuanfeng Ji$^{1,2\boldsymbol{\ddag}}$, Dan Lin$^{3\boldsymbol{\ddag}}$, Xiyue Wang$^{1\boldsymbol{\ddag}}$, Lu Zhang$^{4}$, Wenhui Zhou$^{5}$, Chongjian Ge$^{2}$, Ruihang Chu$^{6}$, Xiaoli Yang$^{1}$, Junhan Zhao$^{7}$, Junsong Chen$^{2}$, Xiangde Luo$^{1}$, Sen Yang$^{1}$, Jin Fang$^{4}$, Ping Luo$^{*2}$, Ruijiang Li$^{*1}$
\end{spacing}

\vspace{-7mm}
\begin{spacing}{1.4}
\begin{affiliations}
    \item Department of Radiation Oncology, Stanford University School of Medicine, Stanford, CA USA
    \item Department of Computer Science, School of Computing and Data Science, The University of Hong Kong, Hong Kong SAR, China
    \item Department of Population Medicine and Diagnostic Sciences, College of Veterinary Medicine, Cornell University, Ithaca, NY, USA
    \item Department of Radiology, The First Affiliated Hospital of Jinan University, Guangzhou, China
        \item Department of Radiology, Stanford University School of Medicine, Stanford, CA USA
    \item Department of Computer Science and Engineering, The Chinese University of Hong Kong, Hong Kong SAR, China
    \item Department of Biomedical Informatics, Harvard Medical School, Boston, MA, USA
 \\$\boldsymbol{\ddag}$ Contributed Equally
 \\\textbf{*Corresponding author}: Ruijiang Li (rli2@stanford.edu), Ping Luo (pluo@hku.hk)
\end{affiliations}
\end{spacing}

\vspace{-3mm}
\begin{spacing}{1.2}

\end{spacing}

\clearpage

%%% Main Text
%\linenumbers
\Heading{Abstract}
\begin{spacing}{1.4}

\noindent
The scarcity of well-annotated diverse medical images is a major hurdle for developing reliable AI models in healthcare. Substantial technical advances have been made in generative foundation models for natural images. Here we develop `ChexGen', a generative vision-language foundation model that introduces a unified framework for text-, mask-, and bounding box-guided synthesis of chest radiographs. Built upon the latent diffusion transformer architecture, ChexGen was pretrained on the largest curated chest X-ray dataset to date, consisting of 960,000 radiograph-report pairs. ChexGen achieves accurate synthesis of radiographs through expert evaluations and quantitative metrics. We demonstrate the utility of ChexGen for training data augmentation and supervised pretraining, which led to performance improvements across disease classification, detection, and segmentation tasks using a small fraction of training data. Further, our model enables the creation of diverse patient cohorts that enhance model fairness by detecting and mitigating demographic biases. Our study supports the transformative role of generative foundation models in building more accurate, data-efficient, and equitable medical AI systems.

\end{spacing}

\newpage

\begin{spacing}{1.35}
\Heading{Introduction}

\noindent
Generative artificial intelligence (AI) models have the potential to revolutionize medical research and clinical care\cite{moor2023foundation,rao2025multimodal}. These models promise to address critical challenges of medical AI development and applications, from synthetic data generation to advanced diagnostic tools, while improving efficiency and outcomes\cite{wang2024self}. Generative AI can be used to create synthetic patient data that mirrors real-world datasets, increasing data volume while enhancing data diversity\cite{ktena2024generative}. This will help overcome a significant hurdle in training robust and generalizable AI models in healthcare, i.e., the lack of large-scale, expert-annotated datasets due to privacy constraints as well as time-consuming and labor-intensive data annotations\cite{sun2025data}. In addition, generative AI can augment or balance underrepresented demographics in real-world training data, improving fairness of AI models\cite{ktena2024generative}.

\noindent
Substantial technical advances have been made in generative vision-language foundation models in the general machine learning community\cite{ho2020denoising,ramesh2021zero,saharia2022photorealistic,rombach2022high,zhang2023adding}. Methods such as Stable Diffusion\cite{rombach2022high} use large datasets with billions of natural image-text pairs to train generative foundation models. Compared to traditional methods such as generative adversarial networks, these diffusion-based models have demonstrated superior capabilities in synthesizing photorealistic images with precise text-guided control\cite{dhariwal2021diffusion,rombach2022high,croitoru2023diffusion}. 

\noindent
Generative foundation models have significant promise for advancing medicine\cite{chambon2022roentgen,wang2024self,ktena2024generative,carrillo2025generation,sun2025data}. When trained on large-scale paired image-text data, these models can learn the relationships between medical images and their text descriptions\cite{chambon2022roentgen,wang2024self}. This allows for controlled synthesis of medical images by specifying desired properties such as demographic attributes and clinical conditions. Among imaging modalities, chest radiographs are a widely used tool for the diagnosis of a variety of medical conditions. However, current X-ray synthesis methods are constrained by limited training data and suboptimal image fidelity due to the use of conventional network architectures\cite{chambon2022roentgen,wang2024self}. More critically, the utility of generative foundation models for X-ray synthesis remains underexplored and their successful applications in medicine are limited\cite{chambon2022roentgen,wang2024self}.

\noindent
Here, we developed ChexGen, a generative vision-language foundation model for chest radiographs that addresses these critical limitations. ChexGen was designed based on a state-of-the-art latent diffusion transformer framework and was pretrained using the largest curated chest X-ray dataset to date, consisting of approximately 960,000 radiograph-report pairs. In contrast to prior models limited to text-to-image generation\cite{chambon2022roentgen,wang2024self}, ChexGen enables precise spatial control of pathology through annotation-to-image synthesis, allowing for mask- and bounding box-guided generation. We assessed the quality of ChexGen-generated images using quantitative metrics and expert evaluations by board-certified radiologists. Furthermore, we demonstrated the utility of ChexGen across multiple downstream applications, including training data augmentation, supervised pretraining, and model bias detection and mitigation. Through extensive experiments on 9 diverse benchmark datasets, ChexGen surpasses existing foundation models in generating clinically realistic chest radiographs with precise textual and spatial control, while achieving state-of-the-art performance in disease classification, anatomical segmentation, abnormality detection, and severity scoring tasks.

% Here, we developed a generative vision-language foundation model named `ChexGen' for generating chest radiographs, one of the most widely used diagnostic tools in clinical practice. ChexGen is designed based on the latent diffusion framework and was pretrained using a large, diverse chest X-ray dataset consisting of approximately 960,000 radiograph-report pairs. We assessed the quality of ChexGen-generated images using quantitative metrics and expert evaluations by board-certified radiologists. Further, we demonstrated the utility of ChexGen for training data augmentation, supervised pretraining, as well as model bias detection and mitigation. Through extensive experiments, we show that ChexGen surpasses existing state-of-the-art foundation models in generating more clinically realistic chest radiographs with precise text control, while achieving superior performance in diagnosing up to 14 diseases across 9 diverse benchmark datasets.

\Heading{Results}

\begin{figure*}
\centering
\includegraphics[width=1.00\textwidth]{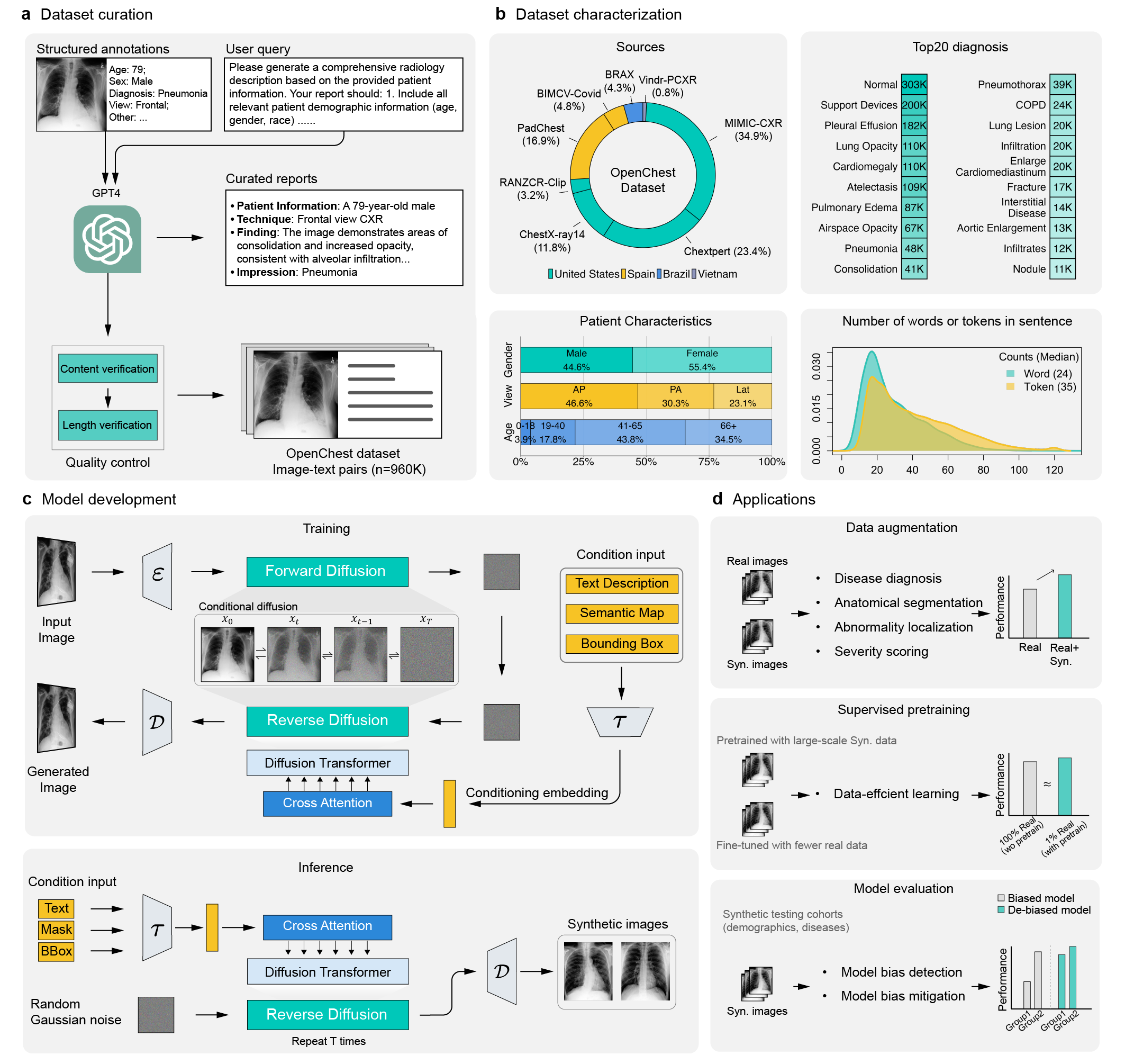}
\end{figure*}
\begin{figure*}
\caption{
\textbf{Data curation, model development and applications.}
\textbf{a,} Dataset curation. We developed OpenChest, a large-scale dataset of 0.96 million chest x-ray image-text pairs aggregated from public datasets. Standardized text reports were generated from structured annotations (e.g., patient demographics and radiological findings) using large language models, with quality control for content and length. \textbf{b}, Dataset characterization. OpenChest aggregates chest radiographs from eight globally distributed cohorts across four continents, encompassing diverse thoracic pathologies, radiographic projections (e.g., anteroposterior, lateral), and patient demographics. Textual reports were quantitatively profiled, with sentence length distributions analyzed through word/token-level density plots. \textbf{c}, Training and inference procedure. ChexGen is a latent diffusion-based generative model, which synthesizes images through conditioning inputs (text descriptions, anatomical masks, or bounding boxes). During training, the model: (1) corrupts input images via forward diffusion with progressive Gaussian noise addition, and (2) reconstructs radiographs through a reverse diffusion process using a transformer architecture with cross-attention to conditioning inputs. Inference generates synthetic images by iteratively denoising Gaussian noise under specified prompts (e.g., textual pathology reports or spatial annotations). \textbf{d}, Downstream applications. (1) Data augmentation enhances thoracic disease diagnosis, anatomical segmentation, abnormality detection, and severity scoring; (2) Supervised pretraining with synthetic data facilitates label-efficient learning and significantly reduces the need for real-world training data. (3) Constructing synthetic test cohorts with specific demographic and disease patterns allows for model bias detection and mitigation. Syn. Synthetic
}
\label{fig:figure1}
\end{figure*}

\heading{Overview of the OpenChest Dataset and ChexGen Model}

\noindent
We developed OpenChest, a large-scale dataset of chest radiographs paired with clinical descriptions. Existing chest radiograph data sets\cite{wang2017chestxray14,irvin2019chexpert,vaya2020bimcv,pham2022vindr,reis2022brax} are often limited to structured annotations (e.g. disease labels and metadata) rather than detailed radiological descriptions. To address this, we developed a large language model-based pipeline that standardizes metadata into radiological descriptions and incorporated quality control measures to ensure accuracy (\textbf{\cref{fig:figure1}a}, \textbf{Methods}).  As a result, we created OpenChest, a dataset comprising approximately 960,000 image-text pairs collected from eight publicly available chest radiograph datasets. OpenChest covers all common chest conditions, includes diverse demographics of patients across four continents, features multiple view positions, along with radiological report descriptions (\textbf{\cref{fig:figure1}b}, \textbf{Extended Data \cref{supptable:dataset_stats}}), making it the largest chest radiograph dataset to date.

\noindent
We then used the OpenChest dataset to train ChexGen, a generative foundation model that synthesizes clinically realistic chest radiographs. ChexGen is built on a latent diffusion framework and comprises three key components: a variational autoencoder (VAE) that transforms images into latent embeddings, a transformer-based diffusion module that iteratively denoises these embeddings to generate images, and condition encoders that handle the processing of condition prompts. During training, the model backbone is trained to denoise latent embeddings that have been deliberately perturbed, with cross-attention mechanisms guided by condition features (\textbf{\cref{fig:figure1}c}, \textbf{Methods}).  During inference, the model performs iterative denoising under given conditions to synthesize chest radiographs (\textbf{\cref{fig:figure1}c}, \textbf{Methods}). 

\noindent
The ChexGen's text-to-image generation capability is achieved through a novel two-stage training strategy (\textbf{Extended Data Figure~\ref{suppfigure:model_arch}}). The first stage focuses on vision-language pretraining, where the model learns coarse alignments between radiographic images and clinical findings using the OpenChest dataset at 256$\times$256 resolution. While this stage provides broad coverage across diverse pathologies and demographics, the automatically generated text descriptions may contain noise or hallucinations. In the second stage, the model is further refined using high-quality, real-world radiology reports from the MIMIC-CXR dataset at 512$\times$512 resolution, enabling fine-grained vision-language alignment and improved synthesis of detailed radiological features. This curriculum learning strategy enables more efficient model optimization compared to traditional single-stage methods~\cite{chen2023pixart,esser2024scaling} (\textbf{Methods}).

\noindent
Building on a robust text-to-image foundation, we further enhanced ChexGen for annotation-to-image generation through the integration of ControlNet\cite{zhang2023adding} and fine-tuning with fine-grained spatial annotations. This enhancement enables precise spatial control over pathology in relation to anatomical structures, allowing disease conditions to appear at any user-specified location in the generated image. This is a unique capability that existing generative models\cite{chambon2022roentgen,wang2024self} are unable to achieve, thereby broadening the scope of controllable medical image synthesis beyond text-only guidance (\textbf{Methods}).

\noindent
We conducted comprehensive evaluations of the generated images through various quantitative metrics and qualitative assessments by board-certified radiologists. Further, we demonstrated the utility of ChexGen in three key applications: synthesizing data for augmentation, supervised pretraining, and model evaluation (bias detection and mitigation) (\textbf{\cref{fig:figure1}d}). When used for data augmentation, ChexGen led to significant improvements in medical image analysis tasks that included disease diagnosis, anatomical segmentation, abnormality detection, and severity scoring (\textbf{\cref{fig:figure3}}). For less common and rare diseases, ChexGen-generated synthetic data can be used for pretraining, which is shown to enhance diagnostic performance (\textbf{\cref{fig:figure5}}). Finally, ChexGen can generate synthetic test cohorts with precisely controlled attributes (e.g., specific disease patterns and demographic characteristics). This provides a powerful approach for detecting and mitigating model bias (\textbf{\cref{fig:figure6}}).

\begin{figure*}
    \centering
    \includegraphics[width=1.00\textwidth]{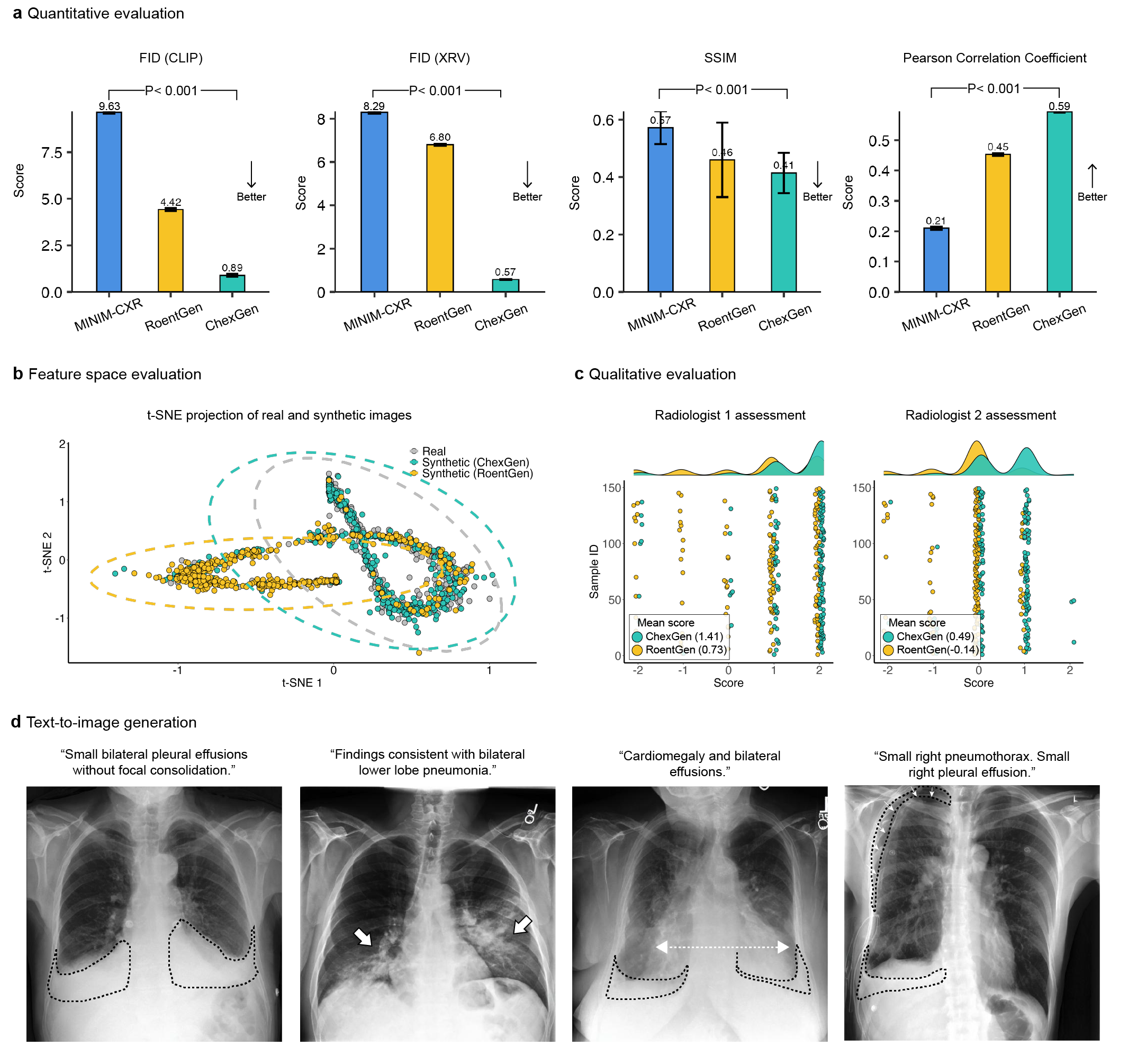}
    \end{figure*}
    \begin{figure*}
    \caption{
    \textbf{Quality evaluations of ChexGen-generated images.}
    We evaluated the quality of chest radiographs synthesized by ChexGen using an independent held-out test set (\textit{n} = 3,500 cases) in MIMIC-CXR that was excluded during model development.
    The synthetic images were generated through text-conditioning on radiological impression reports of the original chest X-rays.
    \textbf{a,} Quantitative evaluation. ChexGen consistently outperformed existing radiograph generation models across different metrics, including Fréchet Inception Distance (FID), Structural Similarity Index Measure (SSIM), and Pearson correlation coefficient. The two-sided Wilcoxon signed-rank test was used to assess the statistical differences between ChexGen and the second-best model (Roentgen). 
    \textbf{b,} Feature space evaluation. t-SNE visualization of deep features extracted by a pre-trained thoracic pathology classifier showed that ChexGen-generated radiographs matched the feature distribution of real chest X-rays more closely compared with RoentGen.
    \textbf{c,} Qualitative evaluation.
    Two board-certified radiologists independently assessed the generated images and corresponding report descriptions using a 5-point scale (-2: complete mismatch, +2: perfect correspondence). Under blinded assessment, ChexGen achieved higher median scores from both raters compared with RoentGen.
    \textbf{d,} Text-to-image generation. Some examples demonstrate the ChexGen's ability to generate condition-specific synthetic radiographs that accurately reflect radiological findings based on text input.
    }
    \label{fig:figure2}
\end{figure*}

\heading{Quality assessment of ChexGen-generated images}

\noindent
We first evaluated the quality of ChexGen-generated images using quantitative and qualitative methods. Synthetic chest radiographs should ideally mimic the distribution of real images (fidelity), capture the full range of variations in the data (diversity), and accurately represent the image features associated with pathological findings (factual correctness). For the quantitative assessment, we used Fréchet inception distance (FID)\cite{heusel2017gans} to evaluate fidelity, multi-scale structural similarity index metric (MS-SSIM)\cite{wang2003multiscale} to measure diversity, and the Pearson correlation score to assess factual correctness. Qualitative assessment was conducted through expert evaluation by board-certified radiologists to verify clinical relevance. We benchmarked ChexGen against state-of-the-art CXR text-to-image generation models, including RoentGen~\cite{chambon2022roentgen} and MINIM-CXR~\cite{wang2024self}.

\hheading{Objective fidelity, diversity, and factual correctness assessment}

\noindent
We assessed fidelity by calculating the FID (lower is better) between synthetic and real CXRs, extracting features from a pretrained general-purpose model (CLIP~\cite{radford2021learning}) and a domain-specific chest radiograph classifier (XRV~\cite{Cohen2022xrv}). ChexGen achieved substantially better fidelity with lower FID scores compared to MINIM-CXR~\cite{wang2024self} and RoentGen~\cite{chambon2022roentgen} (CLIP: 0.89 vs. 9.63 and 4.42; XRV: 0.57 vs. 8.29 and 6.80)  (\textbf{\cref{fig:figure2}a}, \textit{p} < 0.001). These results demonstrate that ChexGen better captures the distribution of real-world chest radiographs.
To evaluate diversity of generated images, we computed the average pairwise SSIM (lower is better) between synthetic images generated from the same prompt but with four different random experiments. ChexGen achieved higher diversity with a lower SSIM score (0.41 ± 0.08) compared to MINIM-CXR (0.57 ± 0.05) and RoentGen (0.46 ± 0.13)  (\textbf{\cref{fig:figure2}a}, \textit{p} < 0.001).
For factual correctness, we computed Pearson correlation coefficients (higher is better) between disease prediction scores derived from synthetic and real images using a pretrained CXR classifier. 
ChexGen achieved a Pearson correlation of 0.59, compared to 0.21 for MINIM-CXR and 0.45 for RoentGen  (\textbf{\cref{fig:figure2}a}, \textit{p} < 0.001). Furthermore, a t-distributed Stochastic Neighbor Embedding (t-SNE) visualization of the two-dimensional feature space (\textbf{\cref{fig:figure2}b}), based on features extracted with the XRV model, revealed that the distribution of ChexGen-generated images aligns with that of real images more closely compared to those generated by RoentGen. This indicates that ChexGen better preserves the underlying data structure of real images.

\hheading{Subjective assessment by radiologists}

\noindent
To assess the clinical authenticity of the synthetic images, two board-certified radiologists (L.Z. and J.F. with 10 and 20 years of experience, respectively) independently evaluated the generated images and their corresponding report descriptions for 150 randomly selected cases from the MIMIC-CXR test set. Using a 5-point scale ranging from -2 (complete mismatch) to 2 (perfect match with radiological findings, as detailed in the Methods section), ChexGen achieved mean rating scores of 1.41 ± 1.01 and 0.49 ± 0.70, while RoentGen scored 0.73 ± 1.31 and -0.14 ± 1.01 for the two radiologists (\textit{p} < 0.001). The score distribution clearly shifted toward higher ratings for ChexGen compared to RoentGen, with a higher concentration of scores in the 1–2 range (\textbf{\cref{fig:figure2}c}). 
Moreover, as demonstrated in \textbf{\cref{fig:figure2}d}, our model can generate chest radiographs across various pathological conditions, including cases with multiple concurrent findings, confirming its ability to generate clinically accurate chest radiographs. Additional examples of synthetic images for different pathological conditions are shown in \textbf{Extended Data Figure~\ref{suppfigure:cond_gen_cls_example}}.

\begin{figure*}
    \centering
    \includegraphics[width=1.00\textwidth]{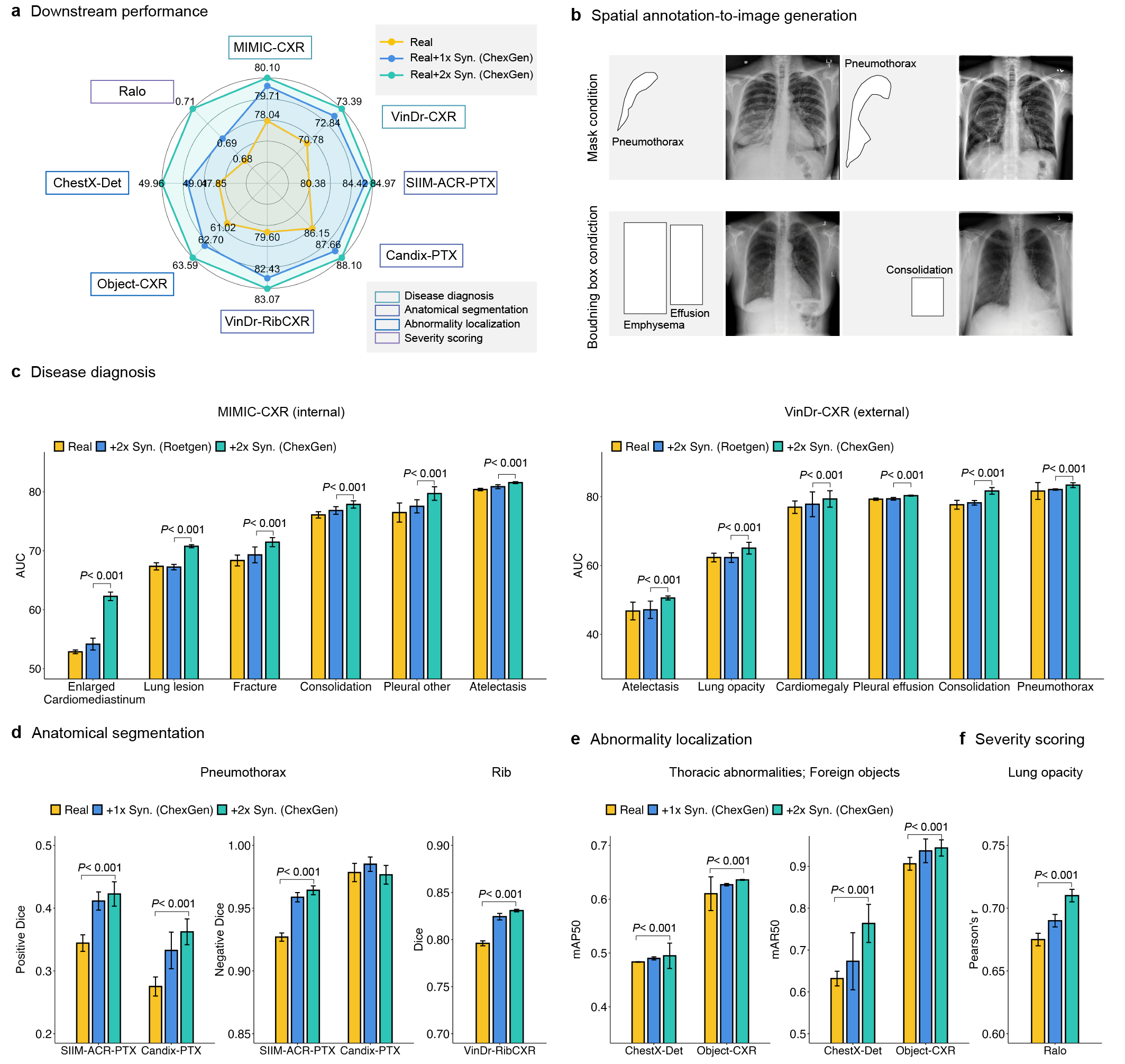}
    \end{figure*}
    \begin{figure*}
    \caption{
    \textbf{Data augmentation using synthetic data.}
    \textbf{a,} 
    Overall performance comparison of models trained on real data versus real+ChexGen synthetic data (1$\times$ and 2$\times$ scaling) across four downstream tasks (disease diagnosis, anatomical segmentation, abnormality localization, and severity scoring) in eight datasets. Synthetic data augmentation demonstrates consistent improvements.
    \textbf{b,} Spatial annotation-to-image examples. Top; Pneumothorax images guided by pneumothorax-specific segmentation mask; CXRs using bounding box condition: emphysema with pleural effusion, and consolidation.
    \textbf{c,} Disease diagnosis performance. ChexGen-augmented models significantly outperform real-data-only and Roetgen-based data augmentation on MIMIC-CXR and independent VinDr-CXR test sets. 
    \textbf{d,} Anatomical segmentation performance: ChexGen-augmented models show consistent improvements over real-data baselines for pneumothorax segmentation in SIIM-ACR-PTX and independent Candix-PTX datasets, with similar performance gains for rib structute segmentation in the VinDr-RibCXR dataset.
    \textbf{e-f,} Abnormality localization and severity scoring performance.
    ChexGen-augmented models demonstrate consistent improvements over real-data baselines in both abnormality localization (foreign objects: Object-CXR dataset; thoracic abnormalities: CheX-Det10 dataset) and lung opacity severity scoring in the RALO dataset.
    In all tasks, model performance increases with the scale of synthetic data augmentation.
    Statistical significance was assessed using the two-sided Wilcoxon signed-rank test, comparing the performance of ChexGen-augmented models against both baseline models or the second-best performing approaches across all experiments.
    }
    \label{fig:figure3}
\end{figure*} 

\heading{Synthesizing data for training data augmentation}

\noindent
We have demonstrated that the ChexGen model generates realistic, diverse chest radiographs that accurately represent pathological features and capture the underlying distributions of real data. These synthetic images can be used to augment training data, which provides a potential solution to the problem of data scarcity and data imbalance frequently encountered in training medical AI models.

\noindent
We evaluated ChexGen's utility for data augmentation in four medical imaging tasks using eight datasets, which include: multi-label disease classification (MIMIC-CXR\cite{johnson2019mimiccxr}, VinDr-CXR~\cite{pham2022vindr}), anatomical segmentation (pneumothorax segmentation on SIIM-ACR-PTX~\cite{siim} and Candix-PTX~\cite{nguyen2021vindr}, rib segmentation on VinDr-RibCXR~\cite{nguyen2021vindr}), abnormality detection (foreign object detection on Object-CXR~\cite{kufel2023chest}, lesion detection on ChestX-Det~\cite{liu2020chestxdet10}), and severity scoring (lung opacity severity regression on RALO~\cite{cohen2021radiographic}). 

\noindent
For each task, we employed task-specific conditional inputs to generate synthetic images (\textbf{\cref{fig:figure3}b}). For example, radiology report impressions served as prompts for classification tasks, while anatomical masks and bounding box annotations guided image generation for anatomical segmentation and disease detection tasks, respectively. Additional examples are shown in \textbf{Extended Data Figures~\ref{suppfigure:cond_gen_cls_example},~\ref{suppfigure:cond_gen_seg_example},~\ref{suppfigure:cond_gen_det_example_object},~\ref{suppfigure:cond_gen_det_example_object_chex10}}. Each generated image was paired with its corresponding condition annotations, and the resulting synthetic dataset was combined with real data for training.  Further details regarding the experimental design and dataset configurations are provided in \textbf{Extended Data Table~\ref{supptable:downstream_stats}} and the \textbf{Methods} section.

\noindent
Across all tasks, we observed that data augmentation with the synthetic data consistently increased model performance (\textbf{\cref{fig:figure3}a}). Adding more synthetic data can bring incremental benefits that start to plateau beyond twice the amount of the original training data. Detailed performance metrics for all tasks are provided in \textbf{Extended Data Tables~\ref{supptable:performance_classification},~\ref{supptable:performance_segmentation},~\ref{supptable:performance_detection}, and~\ref{supptable:performance_regression}}.

\noindent
In disease classification, data augmentation with ChexGen significantly improved model performance across all 14 pathology classes (with six representative classes shown in \textbf{\cref{fig:figure3}c}). The average AUC across all classes increased from baseline (trained with real data) 78.0\% to 79.7\% with 1$\times$ synthetic data and 80.1\% with 2$\times$ synthetic data (\textit{p} < 0.001), while RoentGen's synthetic data only marginally improved the performance (78.0\% to 78.5\%, \textit{p} = 0.21).
The most notable improvements were achieved for challenging conditions such as enlarged CM and lung lesion, with AUC increasing from baseline 52.9\% to 62.3\% (\textit{p} < 0.001) and 67.4\% to 70.8\% (\textit{p} < 0.001) with 2$\times$ synthetic data by ChexGen.
We further evaluated the MIMIC-CXR-trained model on the independent VinDr-CXR dataset (\textit{n} = 3,000).
Testing on six shared pathology classes showed similar patterns with average AUC increasing from baseline 70.8\% to 73.4\% (\textit{p} < 0.001) with 2$\times$ synthetic data by ChexGen (\textbf{\cref{fig:figure2}c}). 
By contrast, adding synthetic data generated by the RoentGen model did not improve upon the baseline results (70.8\% vs 71.2\%, \textit{p} = 0.45).

\noindent
In anatomical segmentation tasks (\textbf{\cref{fig:figure3}d}), data augmentation with ChexGen-generated images showed significant improvements over the baseline model. Training solely on real data produced a Positive Dice score of 0.34 for pneumothorax segmentation in the SIIM-ACR-PTX dataset (\textit{n} = 12,047).
Data augmentation with an extra 2$\times$ synthetic data increased the Positive Dice score to 0.42 (\textit{p} < 0.001), while the Negative Dice remained high at 0.96. 
The improvement generalized to the independent Candix-PTX dataset (\textit{n} = 19,237), where the Positive Dice score increased from 0.28 to 0.36 (\textit{p} < 0.001). 
For rib segmentation, the mean Dice coefficient increased from 0.80 to 0.83 (\textit{p} < 0.001) with 2$\times$ synthetic data on VinDr-RibCXR dataset (\textit{n} = 24,500).

\noindent
Training with synthetic data by ChexGen also improved performance for abnormality detection (\textbf{\cref{fig:figure3}e}). In the ChestX-Det dataset (\textit{n} = 9,000), both mean Average Precision (mAP) and mean average recall (mAR50) at 50\% IoU were significantly improved: mAP increased from 47.9\% to 49.0\% (\textit{p} < 0.001) and mAR50 from 63.2\% to 76.4\% (\textit{p} < 0.001) for detecting six types of disease. Similar improvements were observed on the Object-CXR test dataset (\textit{n} = 1,000) for foreign object detection, where mAP increased from 61.0\% to 63.6\% (\textit{p} < 0.001) and mAR50 from 90.6\% to 94.3\% (\textit{p} < 0.001).
Finally, we observed consistent improvements in the regression task for lung opacity severity scoring on the RALO dataset (\textit{n} = 2,373) (\textbf{\cref{fig:figure3}f}). Training with 2$\times$ synthetic data by ChexGen reduced MAE from 0.65 to 0.58 (\textit{p} < 0.001) and MSE from 0.71 to 0.63 (\textit{p} < 0.001), while improving Pearson correlation coefficient from 0.72 to 0.78 (\textit{p} < 0.001). 

\begin{figure*}
\centering
\includegraphics[width=1.00\textwidth]{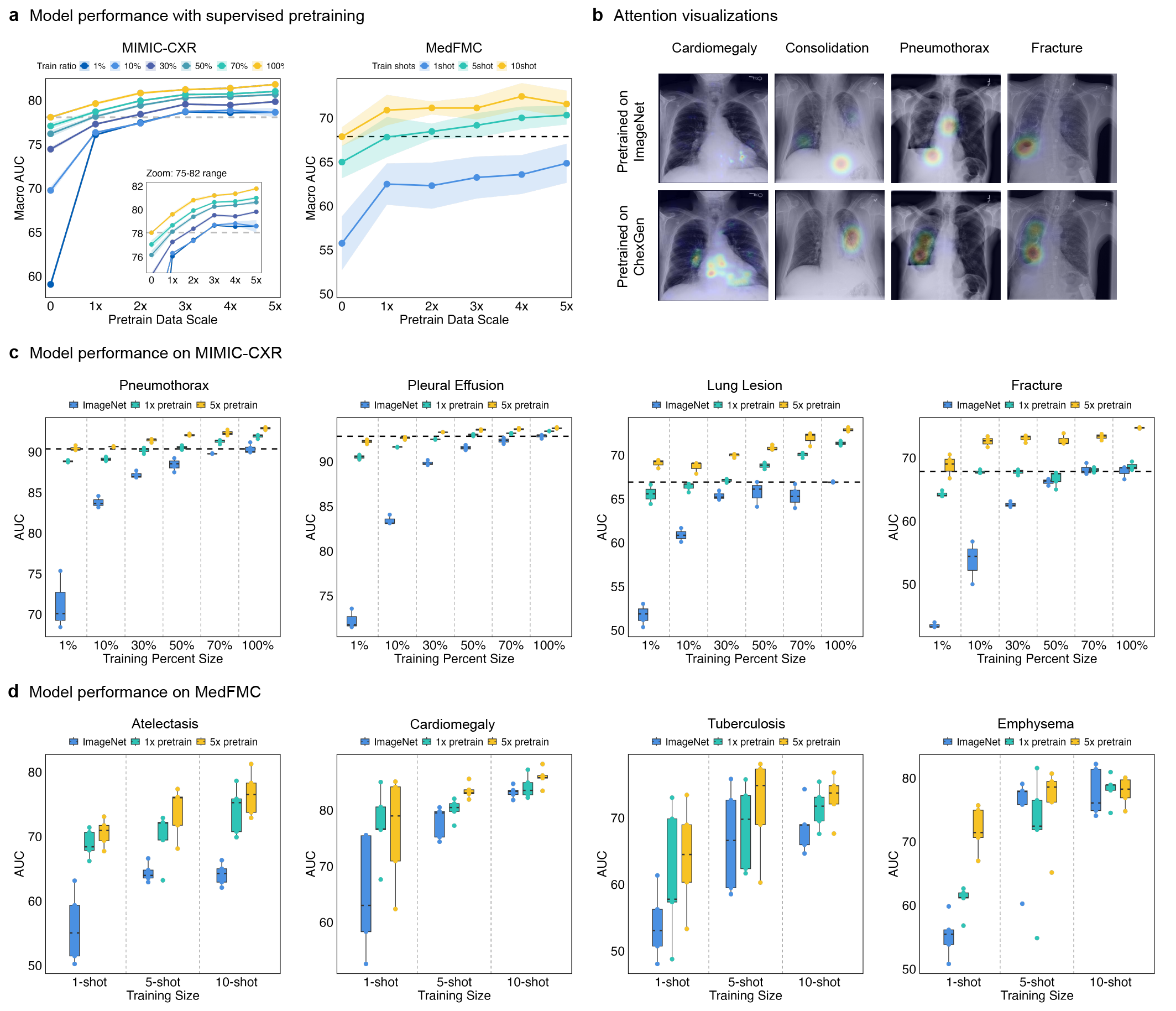}
\end{figure*}
\begin{figure*}
\caption{
\textbf{Supervised pretraining using synthetic data.}
ChexGen-generated data can be used for supervised pretraining, which improves model performance on downstream tasks including disease classification. 
\textbf{a,} Classification performance (macro AUC) for models with varying amounts of ChexGen-synthesized data (from 1$\times$ to 5$\times$ the size of task-specific training data) for pretraining and varying amounts of real data for training. Model performance showed consistent improvements with increasing data scales for pretraining in the MIMIC-CXR dataset. Similar trends were observed in 1-shot, 5-shot, and 10-shot fine-tuning scenarios in the external MedFMC dataset.
\textbf{b,} Class activation maps (CAMs) comparing models pretrained on ImageNet (upper row) with those pretrained on ChexGen data (lower row). Models using ChexGen data show more precise and focused feature localization in pathologically relevant regions.
\textbf{c,} Classification performance (AUC) for models pretrained with ImageNet and varying amounts of ChexGen-synthesized data  (1$\times$ and 5$\times$ the size of task-specific training data), with increasing amount of training data (1-100\%) in the MIMIC-CXR dataset.
\textbf{d,} Few-shot classification performance (AUC) for models pretrained with ImageNet and varying amounts of ChexGen-synthesized data  (1$\times$ and 5$\times$ the size of task-specific training data), with training size from 1-shot to 10-shot in the MedFMC dataset.
Supervised pretraining with ChexGen data consistently demonstrates superior performance over ImageNet pretraining across various training scenarios, ranging from limited data settings to full dataset configurations. 
}
\label{fig:figure5}
\end{figure*}

% \begin{figure*}
% \centering
% \includegraphics[width=1.00\textwidth]{resources/data/figure5.png}
% \caption{
% \textbf{Synthetic data for supervised pretraining.}
% \textbf{a.} 
% Transfer performance, as measured by macro AUC, on the 14-class MIMIC-CRX and 19-class MedFMC datasets using models pretrained on varying volumes of ChexGen-synthesized data. The dashed line indicates the performance of models initialized with ImageNet weights and tuned on the full training set, without synthetic pretraining.
% %
% \textbf{b.} Representative class activation maps (CAMs) illustrate that models pretrained on ChexGen data (bottom) more accurately localize disease-relevant regions compared to the ImageNet-pretrained baseline (top).
% %
% \textbf{c.} On the MIMIC-CXR dataset, classification AUC for diseases such as pneumothorax, pleural effusion, lung lesion, and fracture improves with increasing amounts of ChexGen-synthesized pretraining data (1$\times$ and 5$\times$ scales) across training fractions from 1$\%$ to 100$\%$.
% %
% \textbf{d.} Similarly, on the MedFMC dataset, enhanced classification performance (AUC) for conditions including atelectasis, consolidation, tuberculosis, and emphysema is observed when models are pretrained with ChexGen data under few-shot setting (from 1-shot to 10-shot).}
% \label{fig:figure5}
% \end{figure*}
\heading{Synthesizing data for supervised pretraining}

\noindent
While data augmentation provides a promising solution to the data scarcity problem, simply merging synthetic data with real data for training may not always lead to the optimal results \cite{shumailov2024ai,seddik2024bad}. Here, we propose an alternative strategy to tackle this problem by supervised pretraining on large-scale synthetic data followed by fine-tuning on limited real data. Specifically, we generated synthetic chest radiographs using ChexGen guided by impression sections from the MIMIC-CXR dataset and then pretrained models using disease labels from the original dataset. Fine-tuning experiments were conducted on both MIMIC-CXR and external VinDr-CXR datasets using varying proportions of real training data (ranging from 1\% to 100\%), and on the MedFMC dataset for the most challenging few-shot learning tasks (1-shot and 10-shot). We systematically assessed the impact of varying synthetic data scale (1$\times$ to 5$\times$) for pretraining (see \textbf{\cref{fig:figure5}}). Detailed performance metrics for all tasks are summarized in \textbf{Extended Data Figures~\ref{suppfigure:pretrain_mimic},~\ref{suppfigure:pretrain_vindr}, and~\ref{suppfigure:pretrain_medfmc}}.

\noindent
We found that using only 1\% of the real training data in the MIMIC-CXR dataset, the model's macro AUC substantially increased from a baseline of 59.1\% (pretrained on ImageNet) to 76.1\% when pretrained on 1× synthetic data generated by CheXGen (\textbf{\cref{fig:figure5}a}). As the scale of pretraining data increased from 1× to 5×, model performance consistently improved across all fine-tuning ratios. Remarkably, when pretrained with 5× synthetic data, models fine-tuned on only 1\% of real data achieved a macro AUC of 78.6\%, outperforming the ImageNet-pretrained model trained on the full dataset (78.1\%, as indicated by the dashed line). When more real data is used for fine-tuning, models pretrained on 5$\times$ ChexGen-generated synthetic data achieved even better performance, reaching up to 81.8\% macro AUC with the full dataset.

\noindent
For specific diseases, supervised pretraining with ChexGen synthetic data consistently outperformed ImageNet pretraining across fine-tuning ratios (\textbf{\cref{fig:figure5}c}), with the largest gains achieved in the least amount of training data. For instance, with only 1\% training data, AUC improved from 71.3\% to 88.8\% for pneumothorax, from 72.3\% to 90.5\% for pleural effusion, from 51.7\% to 65.5\% for lung lesions, and from 43.5\% to 64.2\% for fracture classification (\textit{p} < 0.001 for all comparisons). Similar results were obtained for 15-disease classification on the external VinDr-CXR dataset (\textbf{Extended Data Figure~\ref{suppfigure:pretrain_vindr}}).

\noindent
We then tested the models in the few-shot setting for 19-disease classification in the external MedFMC dataset (\textit{n} = 10,000) (\textbf{\cref{fig:figure5}a, right; \textbf{\cref{fig:figure5}d}}). Consistently, we observed that supervised pretraining using ChexGen-generated synthetic data improved model performance, with increasing data scale used for pretraining. In 1-shot experiments, macro AUC increased from 55.7\% (ImageNet pretraining) to 62.5\% with 1$\times$ synthetic data and further to 64.8\% with 5$\times$ synthetic data by ChexGen. Similar trends were observed in 10-shot settings, with macro AUC rising from 67.8\% (ImageNet pretraining) to 70.9\% with 1$\times$ synthetic data and reaching 71.6\% with 5$\times$ synthetic data by ChexGen. For diagnosing specific diseases, in the 1-shot scenario, AUC improved from 55.8\% to 70.5\% for atelectasis and from 65.0\% to 76.3\% for cardiomegaly, while in the 10-shot scenario, AUC increased from 53.9\% to 64.1\% for tuberculosis and from 55.2\% to 71.9\% for emphysema. All differences were statistically significant (\textit{p} < 0.001, two-tailed paired t-test).

\noindent
To further validate the effectiveness of synthetic data for pretraining, we visualized the model's high-attention regions using the established Grad-CAM method\cite{selvaraju2017grad}. While ImageNet-pretrained models exhibited diffuse, non-specific attention, those pretrained with ChexGen's synthetic data consistently focused on disease-relevant anatomical regions. As demonstrated by examples of cardiomely, consolidation, pneumothorax, and fracture, the network effectively concentrates its focus on clinically pertinent areas across diverse conditions (\textbf{\cref{fig:figure5}b}). 

\begin{figure*}
\centering
\includegraphics[width=1.00\textwidth]{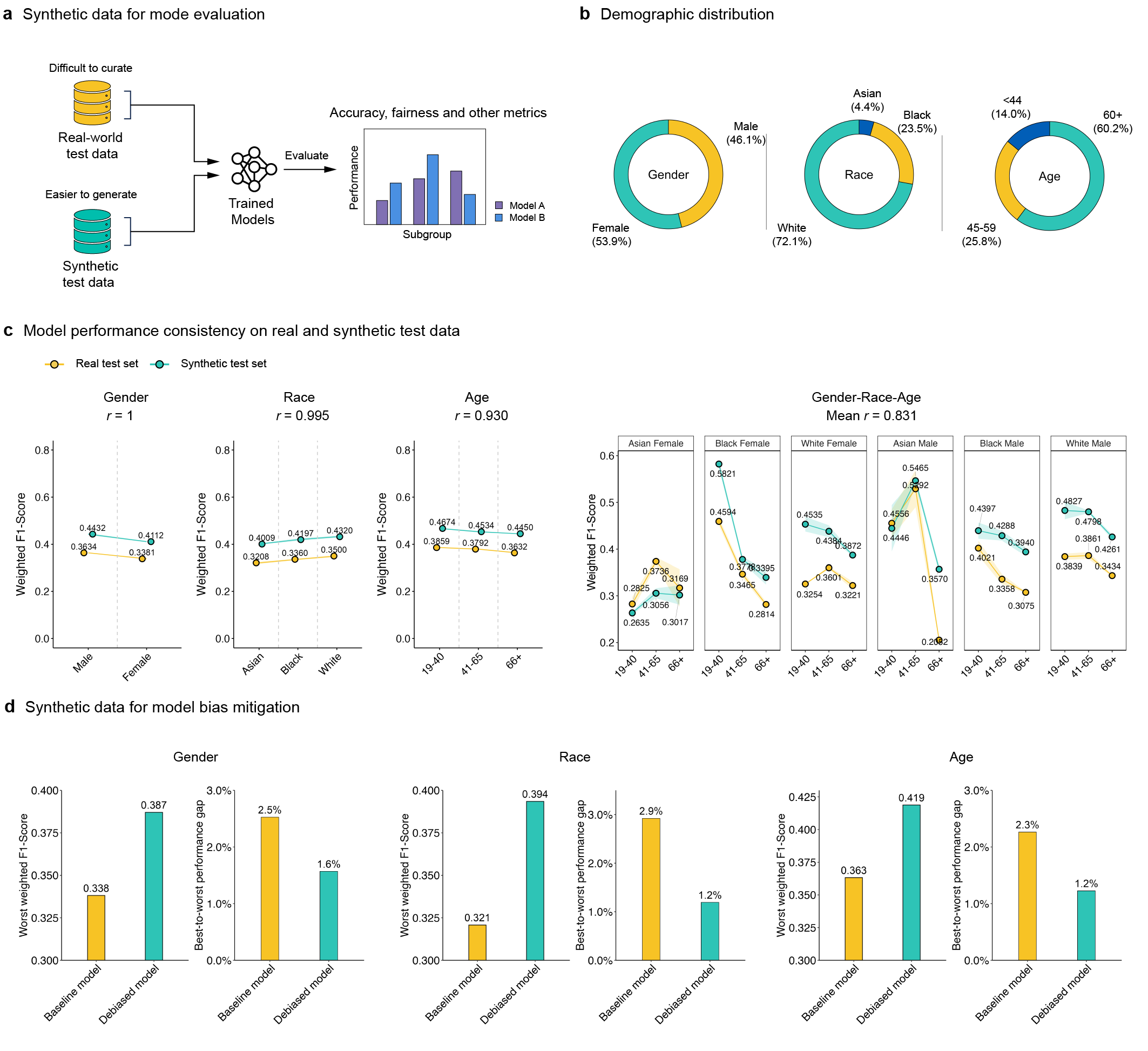}
\end{figure*}
\begin{figure*}
\caption{
\textbf{Bias detection and mitigation using synthetic data.}
ChexGen-generated synthetic data can used to supplement real-world data for model evaluation.
\textbf{a,} Conceptual framework to illustrate that ChexGen-generated data can be utilized as test data to evaluate model performance (e.g., accuracy, fairness), offering an alternative to costly and difficult-to-obtain real-world data.
\textbf{b,} Demographic distribution of the MIMIC dataset. The inherent data imbalances among different demographic groups (e.g., gender, race, and age) could lead to model biases.
\textbf{c,} Synthetic data for model bias detection.
CheXGen is able to generate synthetic test data that matches the demographic composition of real data. Model biases revealed by synthetic test data closely mirror those in real test data, as evidenced by the high correlation coefficients in performance metrics across demographic subgroups.
\textbf{d,} Synthetic data for model bias mitigation.
Once biases are detected, synthetic data can be utilized to enrich underrepresented subgroups, thereby reducing performance gaps between different demographic groups and effectively improving algorithm fairness. 
}
\label{fig:figure6}
\end{figure*}

\heading{Synthesizing data for model bias detection and mitigation}

\noindent
The deployment of AI systems in clinical applications requires rigorous validation to ensure generalizability, reliability, and fairness. Current paradigms require large-scale expert curated evaluation datasets. However, this approach has significant limitations including restricted data access due to privacy regulations, as well as resource- and labor-intensive data collection and annotations. In addition, there is also bias associated with real-world data itself.

\noindent
To overcome these challenges, we propose a model evaluation approach that leverages generative foundation model such as ChexGen to efficiently generate synthetic images with precisely controlled characteristics. This methodology enables the creation of synthetic test cohorts with defined attributes, such as specific disease patterns and demographic characteristics, thereby providing a complementary framework for performance evaluation and benchmarking (\textbf{\cref{fig:figure6}a}). We demonstrated the utility of this approach for assessing algorithm fairness, specifically for bias detection and mitigation.

\noindent
There is pronounced demographic imbalances in the MIMIC-CXR training dataset: White people account for 72.1\% of the cohort, while Black and Asian people make up 23.5\% and 4.4\%, and over 60\% of people are aged 66 or older (\textbf{\cref{fig:figure6}b}). Consequently, the trained model exhibited notable performance disparities across various demographic dimensions, including gender, age, racial subgroups, as well as their combinations, on the real testing dataset (\textbf{\cref{fig:figure6}c}). 
%
% To validate whether synthetic cohorts can reliably detect demographic bias, we first generated synthetic test data with a demographic distribution matching the real-world dataset. 
% %
% Specifically, for each demographic subgroup, we used the corresponding impression reports from the real dataset, augmented with matched demographic attributes (age, race, gender) to generate an equal number of synthetic images. 
% %
% We then evaluated the model trained on the imbalanced training set using both real and synthetic test cohorts.
To validate synthetic cohorts for bias detection, we generated matched test data using real impression reports with corresponding demographic attributes, and evaluated a model trained on the imbalanced training set using both real and synthetic cohorts.
For instance, in the gender-specific analysis, the model demonstrated the lower performance in female subgroup in both real and synthetic data. In the race-specific analysis, the model showed higher performance in the White group compared to other racial groups. In the age-specific analysis, individuals aged 66 and older consistently showed the lowest performance in both datasets. Furthermore, in the combined age–race analysis, the model exhibited the worst performance among Black people in the older age group (66+), with better results for younger people (aged 19–40). Remarkably, we observed strong correlations in model performance between two real and synthetic test cohorts across gender groups(\textit{r} = 1), race groups (\textit{r} = 0.995), age groups (\textit{r} = 0.930),  and gender–race-age combinations (\textit{r} = 0.831).

\noindent
In addition to detecting model bias, we explored whether synthetic data could be used to further mitigate the observed bias. Using ChexGen, we generated additional training samples for underrepresented demographic subgroups until their representation matched that of the majority groups. Consequently, the augmented training data led to substantial improvements in model fairness across all demographic dimensions. For instance, the F1-Scores increased significantly for the worst group (gender: 0.338 to 0.387, race: 0.321 to 0.394, age: 0.363 to 0.419), while performance gaps between best and worst groups narrowed considerably with a relative reduction of up to 60\% (gender: 2.5\% to 1.6\%, race: 2.9\% to 1.2\%, age: 2.3\% to 1.2\%)  (\textbf{\cref{fig:figure6}d}).  These results highlight the potential of CheXGen for generating synthetic data to enhance model fairness by detecting and mitigating bias. 

\Heading{Discussion}

\noindent
In this study, we curated one of the largest, most diverse radiograph datasets with nearly 1 million chest X-ray images and associated text descriptions. By leveraging a state-of-the-art latent diffusion framework, we developed ChexGen, a generative vision-language foundation model for chest radiographs. Our model achieves accurate text-guided medical image synthesis while maintaining diversity and clinical authenticity. This approach addresses key challenges of AI development in healthcare, particularly the scarcity of diverse and well-annotated medical imaging data.

\noindent
Generative foundation models open the door for a wide range of clinical applications: augmenting training data for rare diseases where real samples are scarce; creating diverse patient cohorts to mitigate demographic biases; and enabling systematic stress testing of AI models across various clinical scenarios. However, successful applications of generative models in medical image synthesis have been limited. Our study fills this gap by addressing the following unmet needs for (1) high-performing generative model capable of generating clinically realistic images with precise text control; and (2) comprehensive evaluation of synthetic medical images in various applications ranging from training data augmentation to model bias detection and mitigation. 

% compare with other models to show evidence for superior perf.
\noindent
Through a series of expert evaluations by board-certified radiologists and quantitative metrics, we show that our ChexGen model generates more accurate and realistic chest radiographs compared with existing generative foundation models including RoentGen \cite{chambon2022roentgen} and MINIM \cite{wang2024self}. The clinical utility of ChexGen is demonstrated through three important applications in medical imaging. As a data augmentation tool, it consistently enhances model performance by approximately 10\% across various diagnostic tasks, including disease classification, detection, and segmentation. Further, models pretrained with synthetic data demonstrate remarkable data efficiency, as these models finetuned using only 1\% of real data can achieve diagnostic performance similar to those trained on the whole datasets. This is useful for training AI models for diagnosis of less common or rare diseases. Importantly, ChexGen enables systematic evaluation and mitigation of model biases across demographic groups, addressing a critical need in the validation and deployment of clinical AI models. This will be particularly relevant for medical device regulators such as the FDA to rigorously evaluate AI models and systematically assess for demographic bias before they are broadly deployed in the real world.

% To discuss clinical relevance: any results to highlight, eg, challenging diseases to diagnose
\noindent
Our generative model achieves superior performance in disease diagnosis tasks across diverse pathologies detected on chest radiographs. 
The most significant improvement was observed in enlarged cardiomediastinum, a common diagnosis that is often subject to interpretative variability among radiologists, yielding an 18\% increase in diagnostic accuracy (AUC from 0.529 to 0.623).
Similarly, a notable improvement of 5 \% (AUC from 0.674 to 0.708) was observed in the classification performance for lung lesions, which are significant findings that necessitate workup for early cancer diagnosis.  
Importantly, ChexGen results in a substantial performance boost of 24\% (positive Dice from 0.34 to 0.42) for segmenting pneumothorax and a meaningful increase of 5\% (AUC from 0.683 to 0.715) for diagnosing fractures.  Both conditions represent critical emergencies that require prompt and accurate diagnosis in clinical practice.
The improved performance in these acute conditions is crucial for timely diagnosis and intervention. 
Beyond these specific diagnoses, the superior performance and versatility of ChexGen are broadly generalizable across multiple conditions. 
These combined strengths underscore the tremendous potential of ChexGen as a diagnostic decision support tool, particularly relevant in high-volume or resource-constrained clinical settings.

\noindent
Our ChexGen generative foundation model is distinct from existing foundation models such as CheXagent~\cite{chen2024chexagent} and RAD-DINO~\cite{perez2025exploring} that are focused on image interpretation rather than image generation. 
While useful for diagnosis and report generation, these interpretative models depend heavily on the availability of large-scale diverse data for training purposes. 
ChexGen directly addresses this critical issue by producing clinically realistic synthetic images, thereby enhancing the training and testing of diagnostic models. 
Our approach complements existing systems by increasing data diversity, especially for underrepresented pathologies and demographic groups. 
The future integration between generative and discriminative models will establish a more comprehensive framework, addressing data scarcity and clinical decision-support requirements, ultimately promoting more robust and equitable diagnostic solutions.

\noindent
Our study has several limitations that are worth noting. While ChexGen shows strong performance on relatively common pathologies, its effectiveness for learning representations of various types of rare conditions needs additional validation. The utility of synthetic data in model evaluation, particularly across diverse diseases and patient populations, requires further investigation. Finally, prospective studies in clinical settings would be critical to fully assess the impact of our approach.

\noindent
In conclusion, we develop a generative vision-language foundation model for chest radiographs and demonstrate its applications in training data augmentation, data-efficient learning, as well as bias detection and mitigation. Our results support the transformative role of generative foundation models in building more accurate, data-efficient, and equitable medical AI systems. 

% acknowledgements, data availability, code availability,
% competing interests, author contributions
\noindent\textbf{\large{Acknowledgements}}\\
We acknowledge the MIMIC-CXR, NIH CXR14, CheXpert, PadChest, VinDr-CXR, BIMCV-Covid, Brax, Ranzcr, SIIM-ACR-PTX, Candix, VinDr-RibCXR, Object-CXR, ChestX-Det, RALO, and MedFMC consortia for making their datasets publicly available.

\noindent\textbf{\large{Data Availability}} \\
The MIMIC-CXR dataset is publicly available through PhysioNet (\url{https://physionet.org/content/mimic-cxr/}). 
The NIH CXR14 dataset can be accessed at \url{https://nihcc.app.box.com/v/ChestXray-NIHCC}. 
The CheXpert dataset is available through Stanford University (\url{https://stanfordmlgroup.github.io/competitions/chexpert/}). 
The PadChest dataset can be accessed at \url{http://bimcv.cipf.es/bimcv-projects/padchest/}.
The VinDr-CXR dataset is available at \url{https://vindr.ai/datasets/chest-xray}. 
The BIMCV-Covid dataset can be accessed through \url{http://bimcv.cipf.es/bimcv-projects/bimcv-covid19/}. 
The Brax dataset is available at \url{https://brax.ai/}.
The Ranzcr dataset can be accessed through Kaggle (\url{https://www.kaggle.com/c/ranzcr-clip-catheter-line-classification/}).
The RSNA Pneumothorax dataset is available on Kaggle (\url{https://www.kaggle.com/c/siim-acr-pneumothorax-segmentation/}).
The Candix dataset is available at \url{https://candix.ai/}.
The VinDr-RibCXR dataset can be accessed at \url{https://vindr.ai/datasets/vindr-ribcxr}.
The Object-CXR dataset is available at \url{https://object-cxr.github.io/}.
The ChestX-Det dataset is accessible at \url{https://chestx-det.github.io/}.
The RALO dataset can be found at \url{https://ralo-dataset.org/}.
The MedFMC few-shot learning benchmark is available at \url{https://medfmc.github.io/}.

\noindent\textbf{\large{Code Availability}}\\
The model weights of ChexGen can be requested by researchers credentialed for MIMIC-CXR access. Instructions for obtaining the model weights and the code to implement the diffusion-based medical image generation framework can be accessed via GitHub repository (\url{https://github.com/era-ai-biomed/ChexGen}).

\noindent\textbf{\large{Competing Interests}}\\
The authors declare no competing interests.

\noindent\textbf{\large{Author Contributions}}\\
RL, PL, YJ, DL and XW conceived and designed the study. YJ, DL, LZ were responsible for data acquisition and curation. RC, XY and JC acquired and processed the public datasets. XW, SY, and XL performed the statistical analyses. YJ, DL developed the model architecture. YJ, DL trained and validated the deep learning networks. YJ, XL, and SY implemented quality control of data and algorithms. LZ, WZ and JF were responsible for evaluating the generated results. CG and JZ provided technical support and critical feedback on methodology. YJ, DL, XW, and RL prepared the first draft of the manuscript. RL and PL revised the manuscript. All authors contributed to the preparation of the manuscript and approved the submission for publication.

\end{spacing}
\clearpage

%%%%%%%%%%%%%%%%
%% Figure Legend
%%%%%%%%%%%%%%%%
% \begin{spacing}{1.35}
% \Heading{Figure Legend}
% \input{resources/supplements/0-legend}
% \end{spacing}

%%%%%%%%%%%%%%%%
%% Online Methods
%%%%%%%%%%%%%%%%
\setcounter{figure}{0}
\setcounter{table}{0}

\begin{spacing}{1.35}
\Heading{Methods}\phantomsection\label{sec:methods}
\noindent

\heading{Dataset Curation}

\noindent
In this study we collected chest radiographs from public repositories spanning 2017-2024, adhering strictly to licensing agreements and terms of use. Using a large language model-driven pipeline, we curated the largest known chest radiograph image-text dataset comprising paired radiographs and their corresponding detailed radiological descriptions. 

\noindent
To build a large-scale and diverse dataset, we systematically reviewed and collected eight well-established, publicly available chest radiograph datasets. We specifically focused on datasets with clear institutional origins, well-documented licenses, and structured annotations. The collected datasets can be categorized by their primary functions: general disease classification datasets (NIH ChestX-ray14~\cite{wang2017chestxray14}, Stanford CheXpert~\cite{irvin2019chexpert}, MIMIC-CXR~\cite{johnson2019mimiccxr}, PadChest~\cite{bustos2020padchest}, VinDr-PCXR~\cite{pham2022vindr}, and BRAX~\cite{reis2022brax}), COVID-19 specific datasets (BIMCV-Covid~\cite{vaya2020bimcv}), and a dataset to classify the presence and correct placement of tubes on chest x-rays (RANZCR CLip~\cite{tang2021clip}). These datasets collectively provide comprehensive structured annotations including disease classifications, anatomical landmarks, patient demographics, and imaging parameters.

\noindent
Our data curation framework was designed with two primary objectives: establishing a comprehensive, large-scale dataset and ensuring high-quality medical descriptions for each image. While integrating diverse public chest radiograph datasets from multiple institutions addressed the first objective, a significant challenge emerged: these datasets typically contained only structured annotations (e.g., disease labels and patient metadata) rather than detailed radiological descriptions. To bridge this gap, we developed an automated pipeline for converting these structured annotations into clinically appropriate descriptions. Given the dataset's substantial scale, we implemented a computationally efficient two-stage approach: first utilizing GPT-4~\cite{achiam2023gpt} for the complex task of generating initial medical descriptions, followed by Qwen-2.5~\cite{yang2024qwen2} for systematic quality control.

\noindent
We developed a standardized prompt for GPT-4 specifically designed to process medical metadata including age, sex, disease labels, view position, and clinical parameters. 
This approach converts structured annotations into concise natural language descriptions while preserving essential radiological terminology (\textbf{Extended Data Table~\ref{supptable:openchest_curation}}).
To ensure the quality control of the generated descriptions, we developed a rigorous quality control pipeline utilizing Qwen-2.5, an open-source large language model.
Specifically, Qwen-2.5 was prompted to systematically compare each generated description with its corresponding structured annotations to assess medical accuracy and completeness.
It identified and corrected discrepancies, eliminated redundant information, and supplemented missing details.
This systematic process guarantees both consistency and clinical accuracy (\textbf{Extended Data Table~\ref{supptable:openchest_curation}}).
We iteratively refined our prompts for both report generation and quality control by validating them on a small subset (less than 0.5\%) of the data, and subsequently applied the optimized prompts to the entire dataset.

\begin{figure*}
    \centering
    \includegraphics[width=1.00\textwidth]{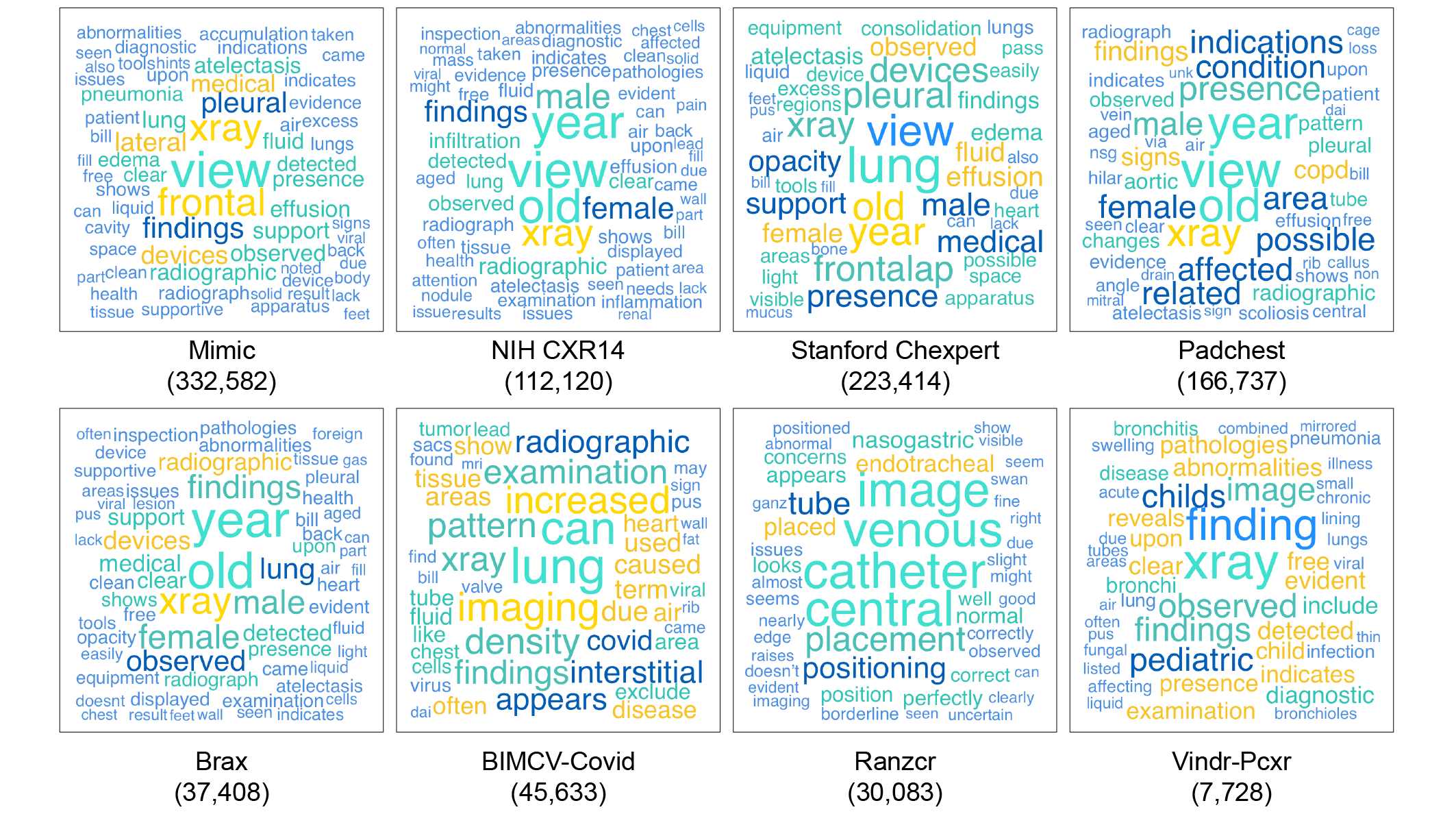}
    \caption{
    \textbf{Word Frequency Visualization of the OpenChest Dataset}. We provide wordcloud visualizations for each source dataset contributing to OpenChest (0.96M pairs), illustrating the distribution and diversity of medical terminology across various institutions and regions. In these visualizations, the size of each term corresponds to its frequency in the standardized radiological descriptions generated by GPT-4. This approach not only underscores the comprehensive coverage of anatomical structures, pathological findings, and clinical observations in our curated dataset but also highlights the distinctive vocabulary patterns unique to each source. 
    }

    \label{suppfigure:word_cloud}
\end{figure*}

\noindent
Through this process, we obtained a final dataset comprising 0.96 million high-quality image-text pairs that establish the correspondence between visual patterns and textual descriptions for various disease conditions. \textbf{Extended Data Table~\ref{supptable:dataset_stats}} provides detailed statistics for each subset, while \textbf{Extended Data Figure~\ref{suppfigure:word_cloud}} illustrates the word frequency distributions across the source datasets.

\heading{ChexGen Model Architecture}

\noindent
ChexGen is built on the latent diffusion model (LDM) framework~\cite{rombach2022high}, which performs diffusion in a compressed latent space rather than directly in pixel space.
This lower-dimensional approach allows LDM to strike an effective balance between generation speed and output quality, allowing efficient synthesis of high-resolution images while preserving visual fidelity, as demonstrated in state-of-the-art natural image generation models such as Stable Diffusion~\cite{rombach2022high} and DALL-E 3~\cite{betker2023improving}.
Aligned with current best practices~\cite{betker2023improving,chen2023pixart}, ChexGen is initially pretrained as a foundation model under a text-to-image paradigm and can subsequently be fine-tuned for spatial annotation-to-image generation.

\hheading{Text-to-Image Generation}

\noindent
ChexGen comprises three primary components: a variational autoencoder (VAE) that encodes images into latent embeddings and reconstructs them, text encoders for processing textual condition prompts, and a diffusion transformer backbone dedicated to denoising (\textbf{Extended Data \cref{suppfigure:model_arch}a}).
We employed the publicly available VAE from Stable Diffusion, which is trained using a combination of perceptual loss and a Kullback-Leibler (KL) divergence objective. 

\noindent
This VAE enables efficient bidirectional mapping between high-dimensional RGB images ($x \in \mathbb{R}^{H \times W \times 3}$) and compressed latent representations ($z = E(x) \in \mathbb{R}^{h \times w \times c}$) through its encoder $E$ and decoder $D$. 
The architecture achieves an 8× spatial compression ratio (i.e., $h = H/8, w = W/8$), significantly reducing computational requirements while preserving essential image features.

\noindent
The text encoder employs a pretrained and frozen T5 model $T_{\text{text}}$, which converts radiological descriptions into semantic embeddings $c^{\text{text}} = T_{\text{text}}(\text{text}) \in \mathbb{R}^{120 \times 4096}$.
The T5 model tokenizes input texts into fixed-length sequences of 120 tokens, using appropriate padding or truncation, and maps each token to a 4096-dimensional embedding.
This longer sequence length (120 tokens vs. 77 tokens in CLIP-based encoders used by previous work~\cite{chambon2022roentgen, wang2024self}) enables our model to process more comprehensive radiology reports, capturing detailed clinical findings and subtle diagnostic nuances that might be truncated in shorter text representations.

\noindent
The diffusion transformer backbone is the core generation component, which employs a 28-block transformer architecture to implement the denoising diffusion process.
Following the DiT-XL/2~\cite{peebles2023scalable} design, each transformer block is equipped with 32 attention heads and operates with 2048-dimensional hidden states.
The architecture alternates between self-attention for processing image features and cross-attention for incorporating text conditions, where the latter attends to projections of $c^{\text{text}}$ to ensure effective text-image alignment.
Notably, diffusion transformer architectures~\cite{peebles2023scalable} adhere to scaling laws. This means that as the volume of training data increases, their performance scales accordingly, in contrast to UNet-based models. 
When combined with our curated, large-scale dataset, this scalability further enhances their capacity for high-quality image synthesis.

\noindent
During training, we process mini-batches of \(M\) image-text pairs \((x_i, \text{text}_i)_{i=1}^M\) through a rigorously designed diffusion pipeline. The VAE encoder maps each input image to its latent representation \(z_i = E(x_i)\), which then undergoes a diffusion process where noise is injected at a randomly sampled timestep \(t \sim U(1,T)\), yielding
\[
z_t = \sqrt{\alpha_t}\, z_i + \sqrt{1-\alpha_t}\, \varepsilon \quad \text{with} \quad \varepsilon \sim \mathcal{N}(0, I).
\]
Simultaneously, the text encoder processes the corresponding description to obtain textual conditional features \(c^{\text{text}}_i = T_{\text{text}}(\text{text}_i)\). Following recent work~\cite{rombach2022high}, we adopt a denoising objective:

\begin{equation}
\mathcal{L}_{\text{DM}} = \frac{1}{M}\sum_{i=1}^M \mathbb{E}_{t,\varepsilon} \left[ \|\varepsilon - \varepsilon_\theta(z_t, t, c^{\text{text}}_i)\|_2^2 \right]
\end{equation}

\noindent
where \(\varepsilon_\theta\) represents the noise prediction network that estimates the noise component conditioned on the noisy latent \(z_t\), the timestep embedding \(t\), and the text features \(c^{\text{text}}_i\). 
The diffusion process follows a linear noise schedule in which \(\beta_t\) increases from \(10^{-4}\) to \(0.02\) over \(T=1000\) timesteps, with the cumulative product 
$\alpha_t := \prod_{s=1}^t (1-\beta_s)$ defining the total noise level at each step. 
To stabilize training and improve text-image alignment, we employ classifier-free guidance with a guidance scale of 4.0 during both training and inference, a strategy that has proven effective in recent text-to-image diffusion models~\cite{ho2022classifier}.

\noindent
At inference, the generation process is initiated with a radiological description \(\text{text}\) as input. Following the standard diffusion sampling procedure, we first sample an initial latent representation \(z_T \sim \mathcal{N}(0, I)\) from a Gaussian distribution. 
The image is then synthesized via an iterative denoising process using the DDIM sampler~\cite{song2020denoising} with 100 refinement steps, under the guidance of the text condition. 
Specifically, at each denoising step, the model computes:

\begin{equation}
z_{t-1} = \sqrt{\alpha_{t-1}} \left[\frac{z_t - \sqrt{1-\alpha_t}\,\varepsilon_\theta(z_t,t,c^{\text{text}})}{\sqrt{\alpha_t}}\right] + \sqrt{1-\alpha_{t-1}}\,\varepsilon_\theta(z_t,t,c^{\text{text}})
\end{equation}

where the predicted noise \(\varepsilon_\theta\) guides the gradual transformation from random noise to a chest radiograph that faithfully reflects the input description.

\hheading{Annotation-to-Image Generation}

\noindent
Although current text-to-image generative models can create images from text prompts, they are limited in controlling the spatial composition of the generated images. Relying solely on text makes it challenging to accurately represent complex anatomical structures and precisely locate lesions.
To overcome this limitation, we enhanced ChexGen with a lightweight ControlNet adapter~\cite{zhang2023adding} that enables finer-grained spatial control during image generation (\textbf{Extended Data \cref{suppfigure:model_arch}b}).
Specifically, the ControlNet module augments standard text-to-image synthesis by accepting spatial annotations (e.g., segmentation masks, lesion bounding boxes) as additional conditioning inputs. 
These inputs allow us to precisely control both pathological features and physiological structures, resulting in images that accurately reflect the desired attributes.
In our approach, spatial annotations are formatted as images and encoded using a VAE encoder to produce latent embeddings: $\mathbf{c}^{\text{anns}} = E(\text{anns})$.

\noindent
The ControlNet adapter processes these embeddings to extract conditioning features, which are then incorporated into the diffusion process through zero-initialized convolutions. 
This mechanism guides the denoising process, ensuring that the final image adheres to the spatial layout specified by the annotations.
Consequently, our approach not only generates images from textual prompts, but also produces image pairs that link each synthesized image with its spatial annotation.
The dual output which can be used to diverse, task-specific datasets, thereby facilitating the development of downstream medical imaging applications.
Notably, the pretrained transformer backbone is kept frozen, and only the lightweight ControlNet adapters are trained. 
This modular approach allows for efficient adaptation to new medical tasks by fine-tuning only a limited number of parameters."

\heading{ChexGen Model Training}

\begin{figure*}
    \centering
    \includegraphics[width=1.00\textwidth]{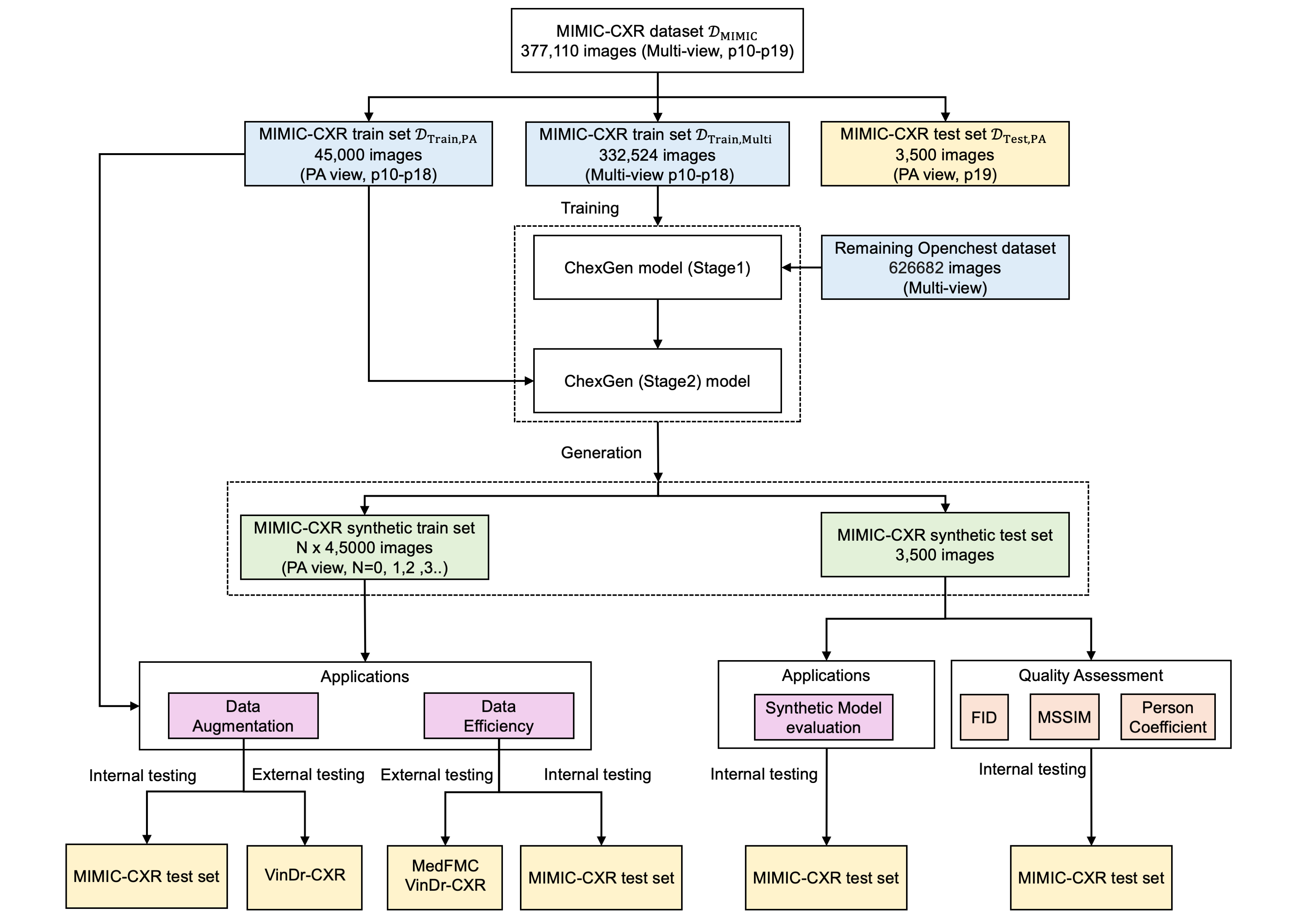}
    \caption{
    \textbf{The datasets used in the ChexGen pertaining and downstream application phases.} (1) Pretraining Phase. MIMIC-CXR (377,110 images) is systematically split into a PA view training set (45K), a multi-view training set (332K), and a PA view test set (3.5K). OpenChest (960K pairs) is constructed using the multi-view training set (332K) and additional datasets (627k pairs).  
    ChexGen follows a two-stage training process.  In Stage I, OpenChest is used for vision-language pretraining. In Stage II, the MIMIC-CXR PA view (45K pairs) is used for alignment. After pertaining, ChestGen can generate N$\times$ synthetic data for both training and test sets in downstream tasks. (2) Downstream Applications. The synthetic data is first evaluated using FID, MSSIM, and Pearson Coefficient metrics for quality assessment. We then apply it in three key applications: data augmentation to enhance model training, data efficiency to improve performance with limited real data, and synthetic model evaluation to assess model reliability on generated samples. For brevity, we omit the detailed statistics of multiple downstream datasets used in various tasks. 
    }

    \label{suppfigure:mimic_dataset_analysis_flowchart}
\end{figure*}

\begin{figure*}
    \centering
    \includegraphics[width=1.00\textwidth]{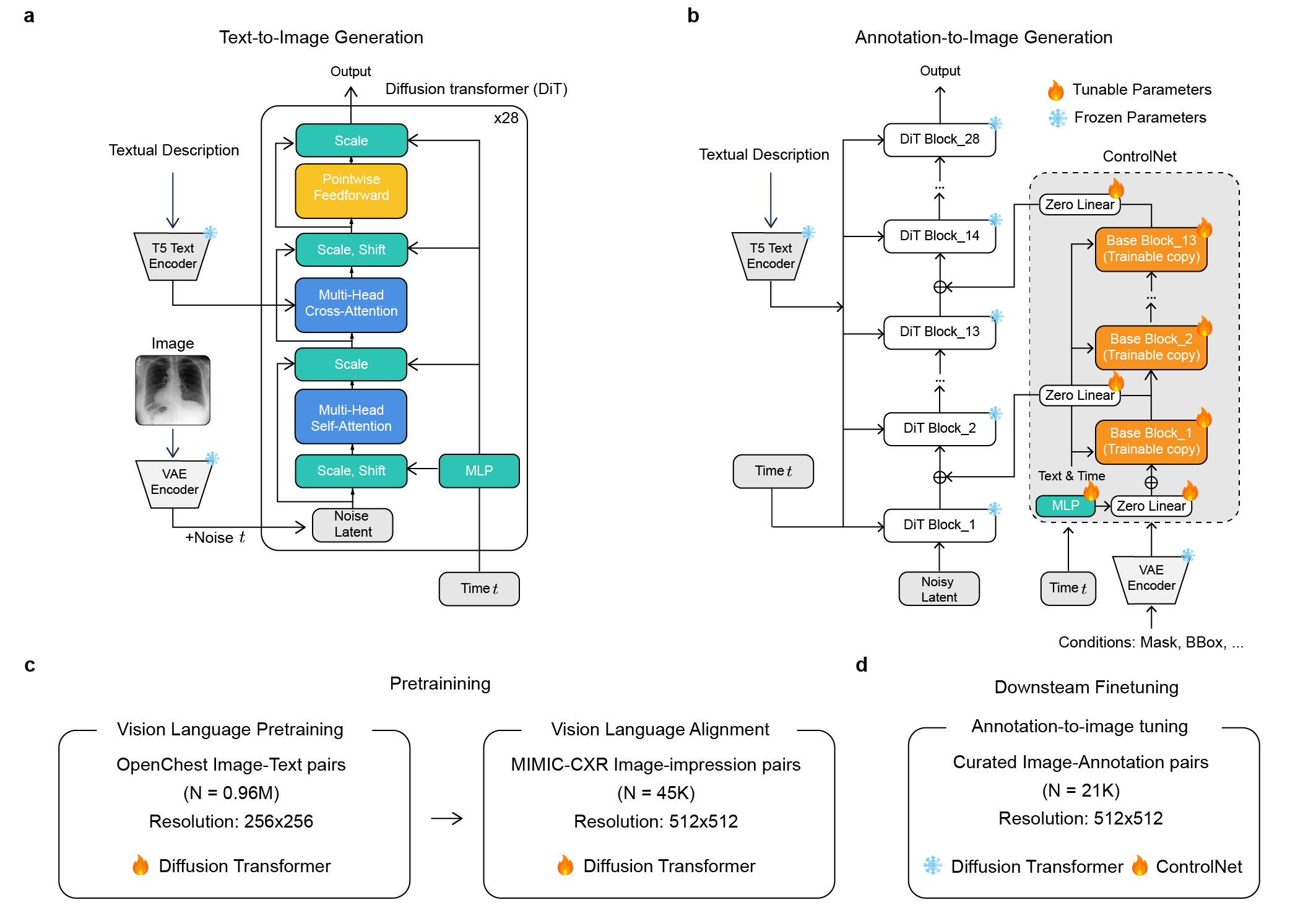}
    \caption{
    \textbf{Overview of the ChexGen model architecture and training strategy.} 
    \textbf{a.} ChexGen model architecture. ChexGen is pretrained under the text-to-image generation paradigm. ChexGen comprises three main components: a frozen T5 text encoder, a frozen Stable Diffusion VAE encoder, and a diffusion transformer. The T5 encoder processes clinical descriptions, while the VAE encoder extracts latent image features. The diffusion transformer then integrates these representations by processing input noise and timesteps through DiT blocks equipped with cross-attention mechanisms, enabling robust text-guided image synthesis.
    \textbf{b.} Enhanced ChexGen architecture for downstream tasks. To achieve spatial annotation-to-image generation, we enhanced ChexGen by integrating a lightweight ControlNet adapter. This adapter processes spatial annotations, such as segmentation masks and bounding boxes, providing additional conditioning signals that facilitate precise control over anatomical structures and pathological findings during image generation. 
    \textbf{c.} ChexGen model pretraining. The pretraining is executed in two stages. Stage I (Vision-Language Pretraining) uses the OpenChest dataset (0.96M image-text pairs) at a resolution of 256$\times$256. In this stage, GPT-4 generated descriptions serve as conditioning signals to establish a fundamental text-image correspondence. Stage II (Vision-Language Alignment) fine-tunes the model using MIMIC-CXR clinical reports (45K pairs) at a higher resolution of 512$\times$512, aligning the model with professional radiological descriptions to enhance clinical fidelity. 
    \textbf{d.} ChexGen model finetuning for spatial annotation-to-image generation. During finetuning, the ControlNet adapter is used to incorporate diverse spatial annotations, such as segmentation masks and bounding boxes, as additional conditioning signals. The model is trained on 21K image-annotation pairs, enabling it to generate images with precise anatomical layouts and accurate pathological representations. }
    \label{suppfigure:model_arch}
\end{figure*}

\begin{figure*}
    \centering
    \includegraphics[width=1.00\textwidth]{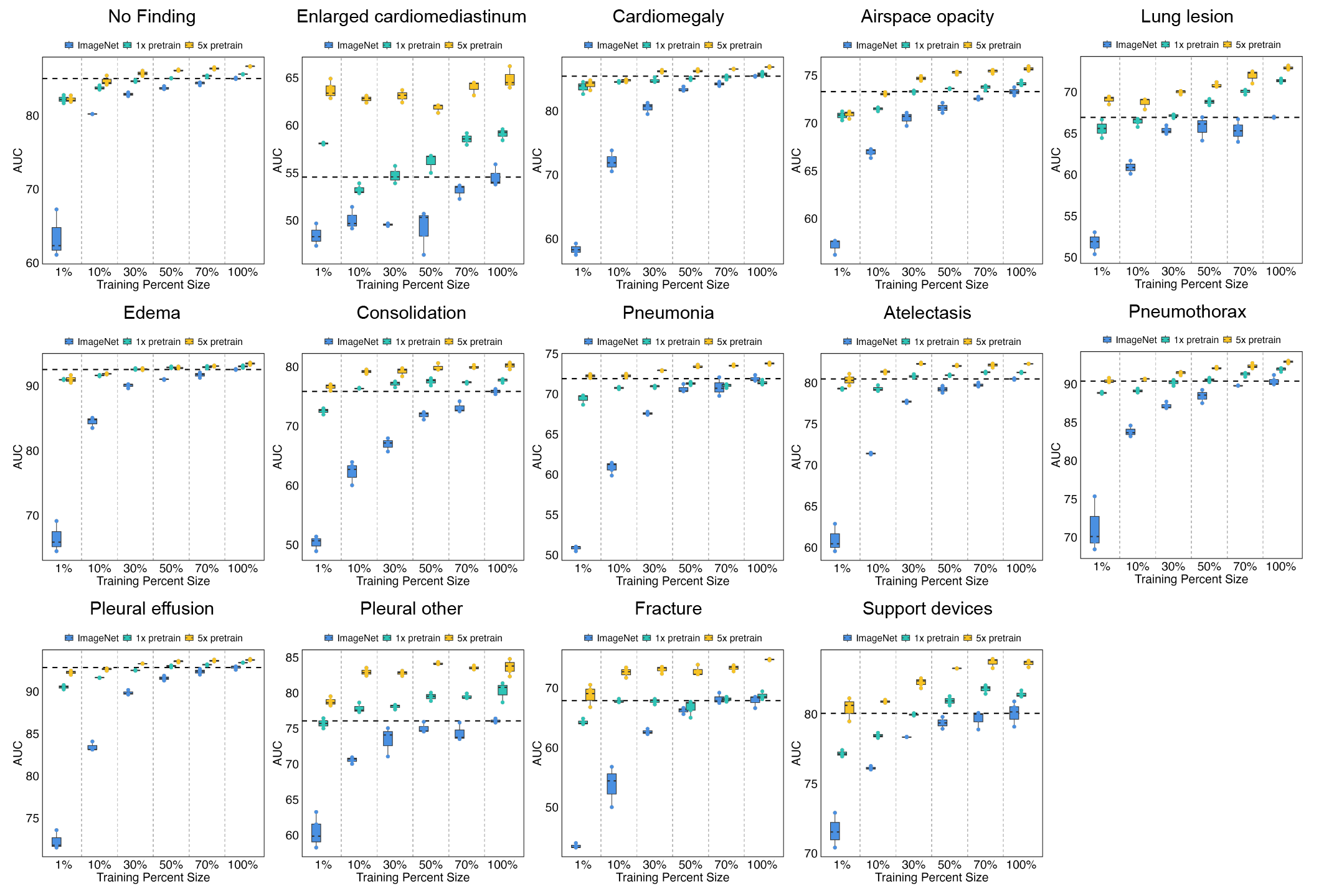}
    \caption{
    \textbf{ChexGen synthetic data used for supervised pretraining in downstream classification tasks with the performance evaluation on the MIMIC-CXR test set}. DenseNet serves as the classification network, and we compare its performance when pretrained on ChexGen synthetic data versus ImageNet.The ChexGen generated data based on the conditions of the MIMIC-CXR training set.
    After pretraining, DenseNet is further trained on the MIMIC-CXR training dataset and evaluated across all 14 disease categories in MIMIC-CXR test set.  The evaluation spans different data regimes (1\%, 10\%, 50\%, and 100\% of training data) to assess data efficiency. Results demonstrate that: (1) ChexGen pretraining consistently outperforms ImageNet pretraining across all disease categories; (2) The performance advantage is particularly pronounced in low-data scenarios (1-10\% training data), where ChexGen achieves comparable results to ImageNet training with full data; (3) The improvements are consistent across both common conditions (e.g., Atelectasis and Cardiomegaly) and rare pathologies, indicating robust feature representations for diverse medical conditions. 
    }
    \label{suppfigure:pretrain_mimic}
\end{figure*}

\begin{figure*}
    \centering
    \includegraphics[width=1.00\textwidth]{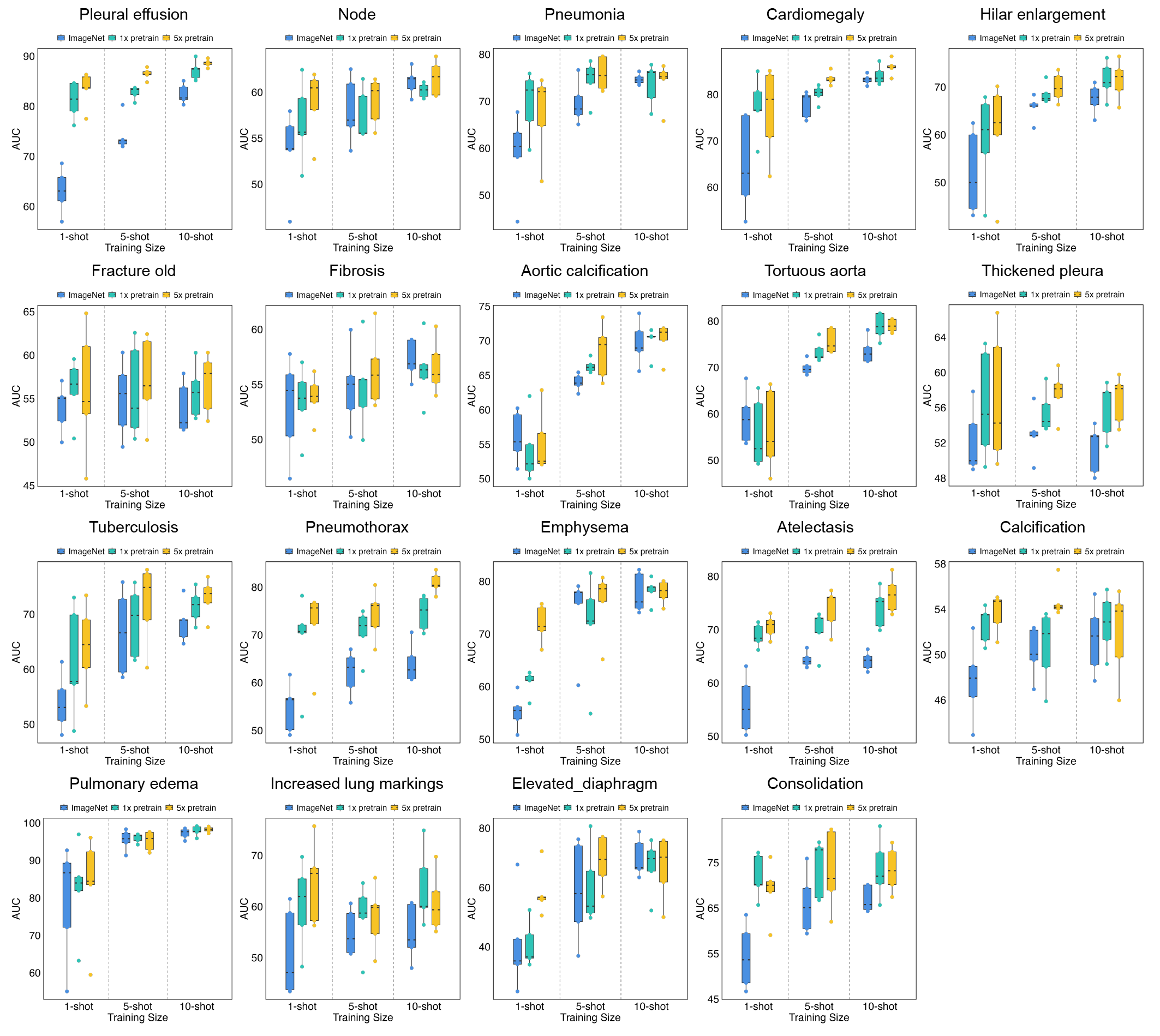}
    \caption{
    \textbf{ChexGen synthetic data used for supervised pretraining in downstream classification tasks with the performance evaluation on the MedFMC test set}.
    DenseNet serves as the classification network, and we compare its performance when pretrained on ChexGen synthetic data versus ImageNet. The ChexGen generated data based on the MIMIC-CXR training set.
    After pretraining, DenseNet is further trained on the MedFMC training dataset and evaluated across all 19 disease categories in MedFMC test set. 
    % We present a systematic analysis of model transferability across all 19 disease categories in MedFMC, an independent external dataset. 
    The evaluation compares ChexGen pretraining (using synthetic data generated by ChexGen) against ImageNet pretraining in few-shot learning scenarios: (1) In 1-shot settings, ChexGen pretraining achieves significant improvements across multiple conditions (e.g., Pneumothorax from 54.8\% to 71.8\%, Pleural Effusion from 63.1\% to 83.4\%); (2) The performance gains scale consistently with shot numbers, showing robust improvements in 5-shot and 10-shot scenarios; (3) The enhanced performance spans diverse pathological categories, from common findings to rare conditions, demonstrating effective knowledge transfer across different medical domains and institutions. 
    }
    \label{suppfigure:pretrain_medfmc}
\end{figure*}

\begin{figure*}
    \centering
    \includegraphics[width=1.00\textwidth]{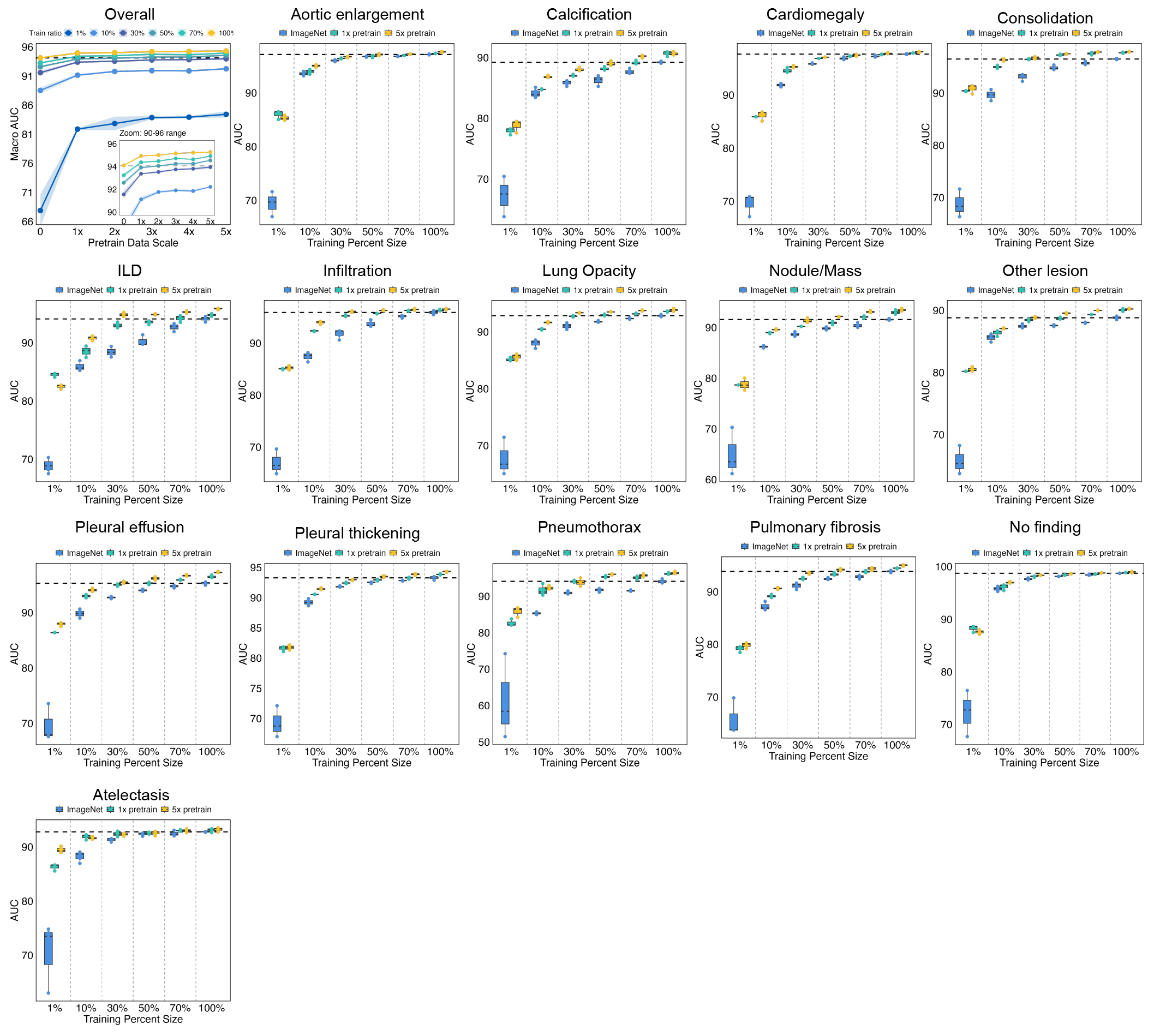}
    \caption{
    \textbf{ChexGen synthetic data used for supervised pretraining in downstream classification tasks with the performance evaluation on the VinDr-CXR test set.} 
    DenseNet serves as the classification network, and we compare its performance when pretrained on ChexGen synthetic data versus ImageNet. The ChexGen generated data based on the MIMIC-CXR training set.
    After pretraining, DenseNet is further trained on the VinDr-CXR training dataset and evaluated across all 15 disease categories in VinDr-CXR test set. 
    We compare performance across different training ratios (1\%, 10\%, 50\%, and 100\%), and observe a clear improvement in classification performance across 15 evaluated disease categories.
    }
    \label{suppfigure:pretrain_vindr}
\end{figure*}

\begin{figure*}
    \centering
    \includegraphics[width=1.00\textwidth]{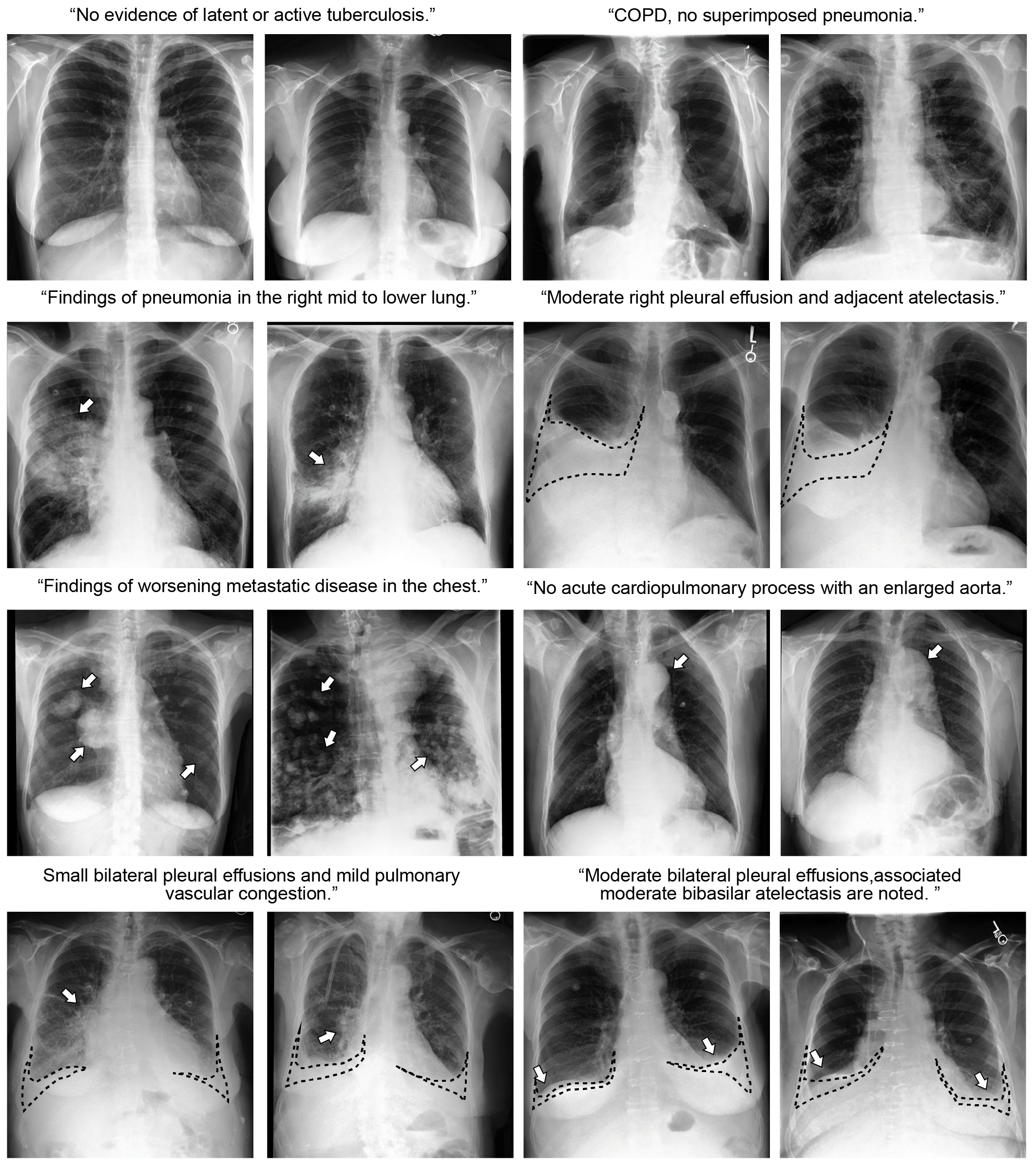}
    \caption{
    \textbf{Examples of text-conditioned CXR generation.} This figure demonstrates ChexGen's ability to generate chest X-ray images conditioned on disease categories using text prompts. The model produces diverse and realistic images that accurately reflect the specified disease patterns, faithfully representing the pathological regions while preserving the natural appearance of lung tissues and overall anatomical structure. 
    }
    \label{suppfigure:cond_gen_cls_example}
\end{figure*}

\begin{figure*}
    \centering
    \includegraphics[width=1.00\textwidth]{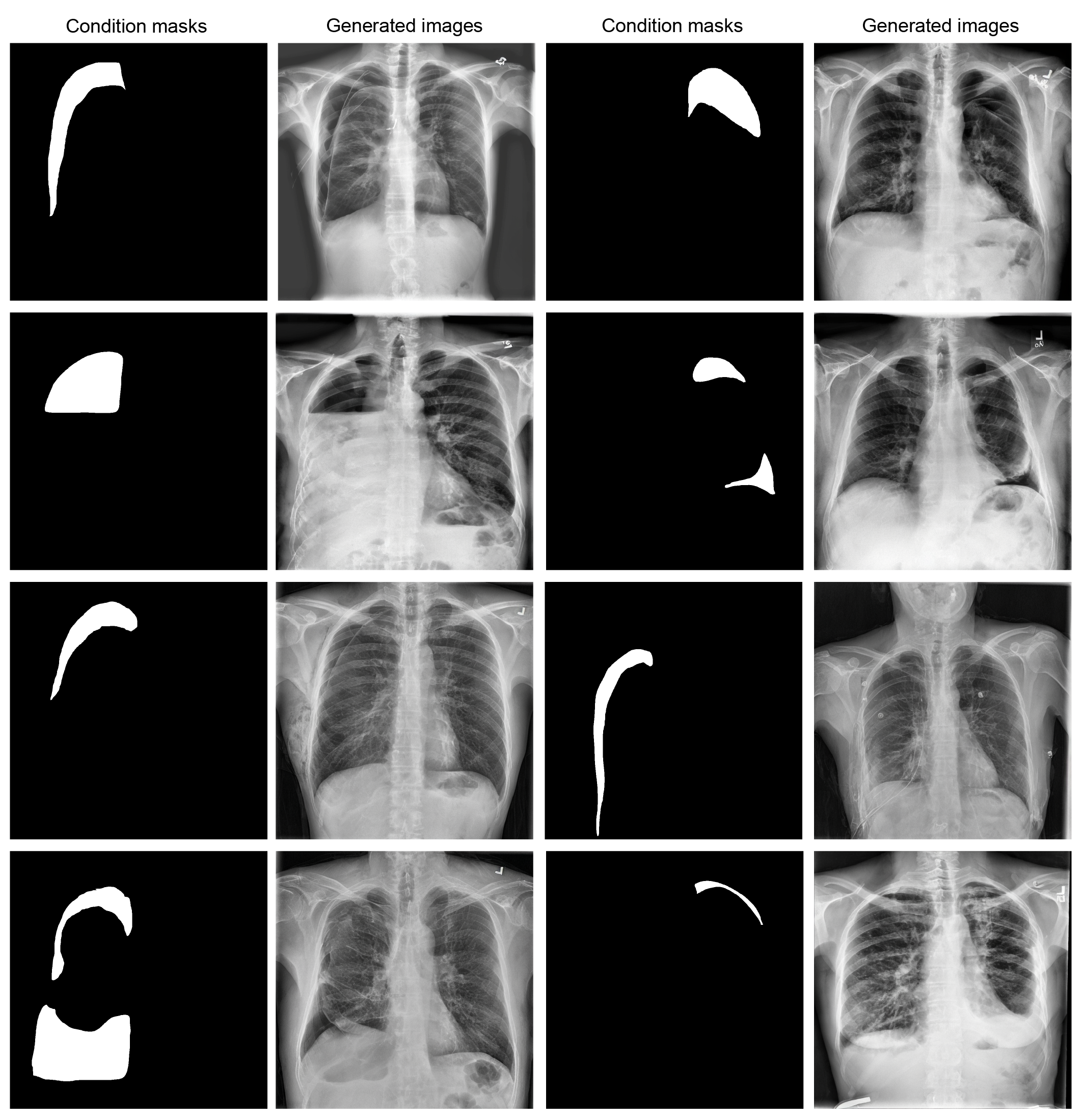}
\caption{
\textbf{Examples of mask-conditioned CXR generation.} This figure demonstrates ChexGen's ability to generate chest X-ray images conditioned on segmentation masks (pneumothorax). Each row displays a set of generated images, with the left-most column showing the input condition and the following three columns presenting the generated samples. 
}

    \label{suppfigure:cond_gen_seg_example}
\end{figure*}

\begin{figure*}
    \centering
    \includegraphics[width=1.00\textwidth]{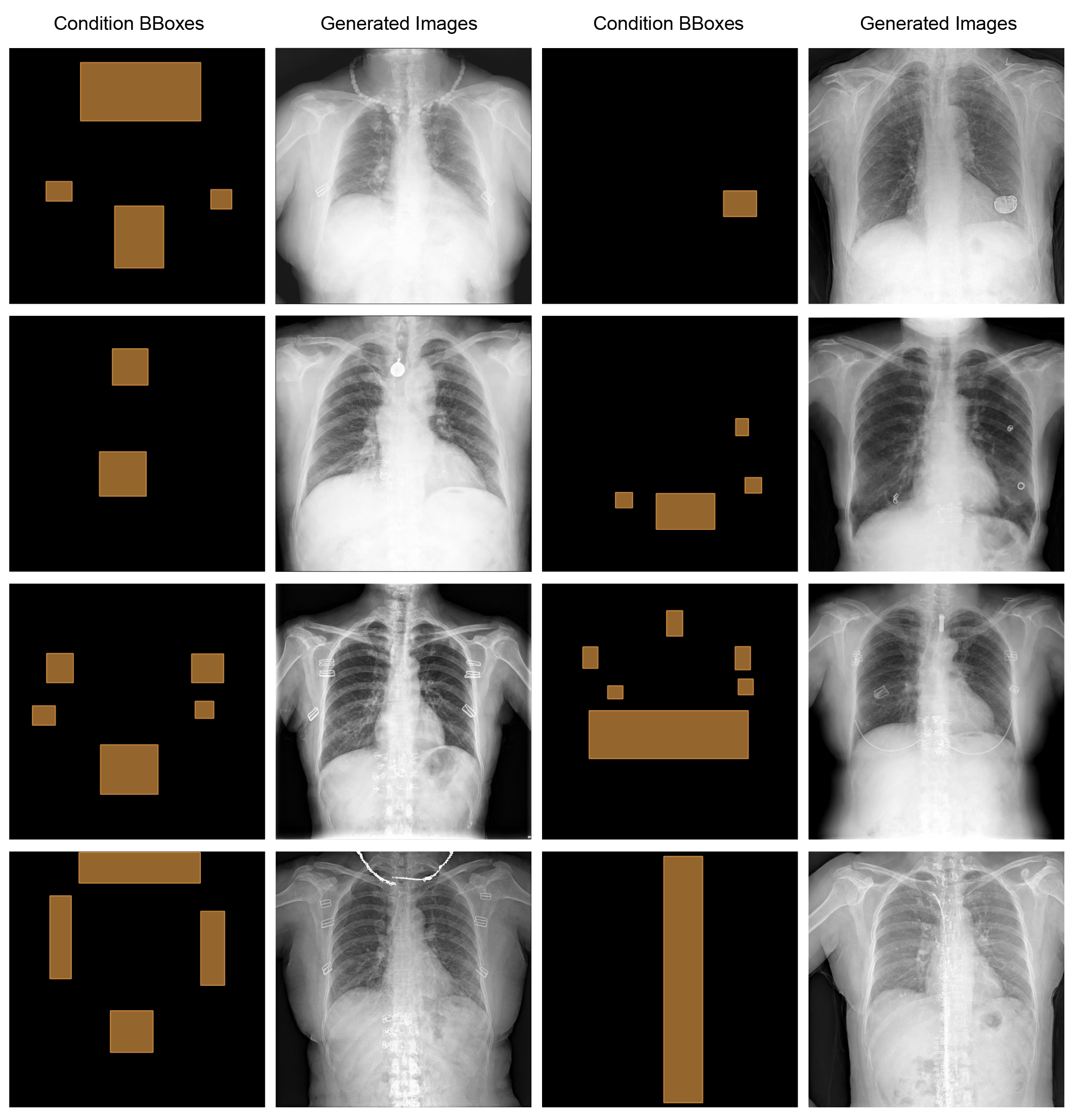}
    \caption{
        \textbf{Examples of bounding box-conditioned CXR generation.} 
        Demonstration of ChexGen's capability to generate chest X-ray images conditioned on foreign object locations using bounding boxes. Each row shows different examples where the left-most column displays the input condition (bounding boxes indicating the locations of foreign objects such as medical devices and implants), followed the generated samples. The model successfully generates diverse yet realistic chest X-ray images that accurately incorporate the foreign objects at the specified locations. 
        }
    \label{suppfigure:cond_gen_det_example_object}
\end{figure*}

\begin{figure*}
    \centering
    \includegraphics[width=1.00\textwidth]{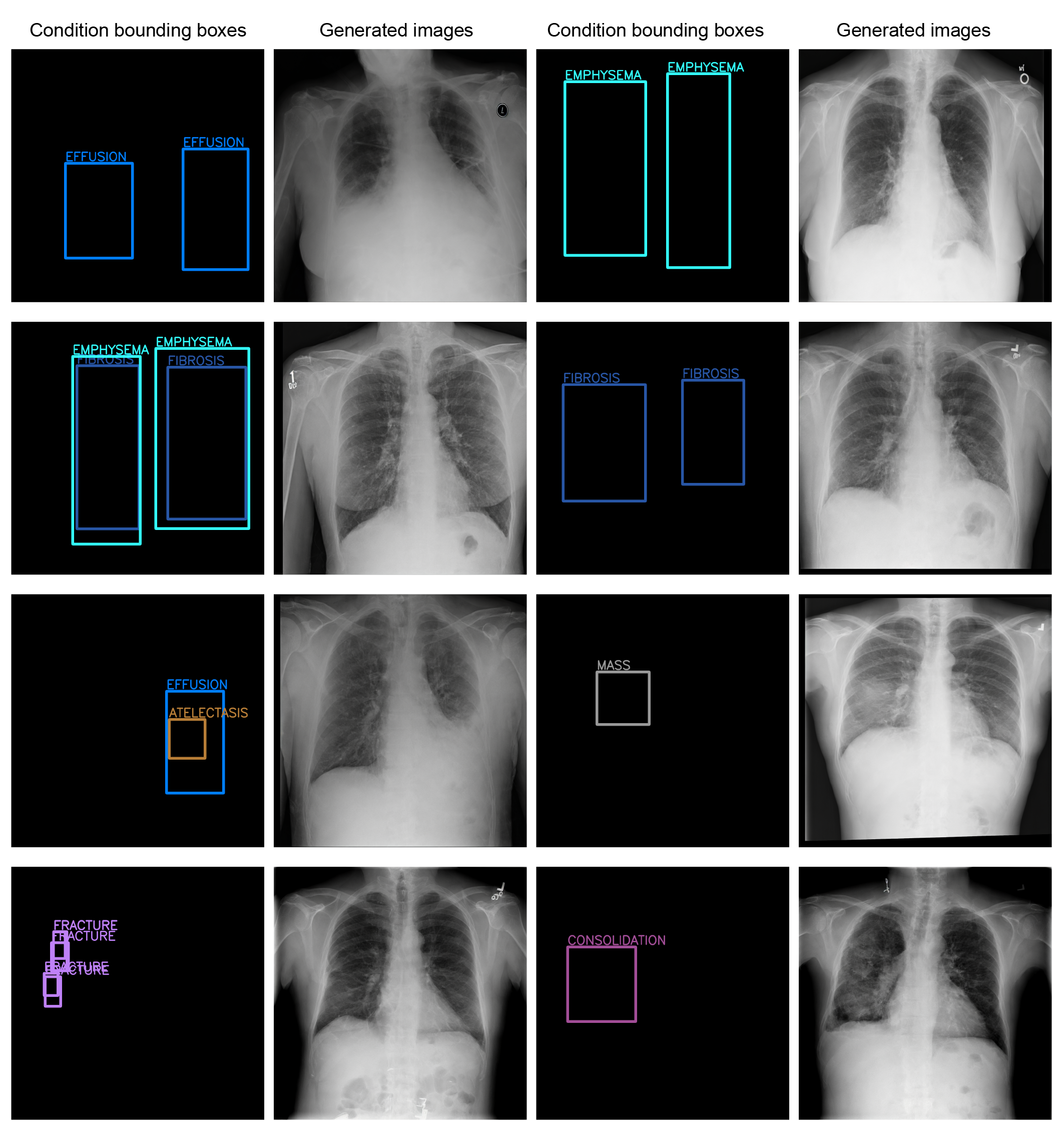}
    \caption{
    \textbf{Examples of pathological bounding box-conditioned CXR generation.} This figure demonstrates ChexGen's capability to generate chest X-ray images conditioned on specific pathology locations using bounding boxes. Each row presents a set of generated images, where the left-most column shows the input condition (bounding boxes with corresponding pathology labels, e.g., effusion, emphysema, consolidation, and fracture), followed by the generated samples. The model successfully produces diverse and realistic chest X-ray images that respect both the spatial constraints (box locations) and the pathological characteristics specified in the input conditions. Different colors of bounding boxes denote different pathology types. 
    }
    \label{suppfigure:cond_gen_det_example_object_chex10}
\end{figure*}

\noindent
ChexGen is pretrained using a two-stage strategy to ensure robust generation capabilities and high quality medical image synthesis.
\textbf{Extended Data Figure ~\ref{suppfigure:mimic_dataset_analysis_flowchart}a} illustrates the comprehensive workflow of ChexGen's training process, from dataset organization through vision-language pretraining and alignment.
In Stage I (Vision-Language Pretraining), we train the model on our curated OpenChest dataset $\mathcal{D}_{\text{OpenChest}}$, which contains 959,206 image--text pairs, at a resolution of 256$\times$256 for 800 epochs. 
This stage enables the model to learn the fundamental correspondence between radiological findings and visual patterns.
Pretraining is performed with a batch size of 192 per GPU across 32 A100 GPUs.
In Stage II {Stage II: Vision-Language Alignment}, we fine tune the model using a carefully curated subset $\mathcal{D}_{\text{Train, PA}}$ of the MIMIC-CXR dataset (version 2.0.0).
We specifically select 45,000 high quality image report pairs from folders p10 to p18, focusing on posterior anterior (PA) view cases. In these cases the impression sections of the reports, written by board certified radiologists, contain standardized clinical findings and interpretations in no more than 120 tokens. 
Stage II is conducted at an increased resolution of 512$\times$512 for 300 epochs, with a reduced batch size of 64 per GPU, which helps the model adapt to authentic radiological descriptions and generate higher resolution images. 
Detailed dataset statistics can be found in \textbf{Extended Data Table~\ref{supptable:dataset_stats}} and a more detailed summary of our training strategy is provided in \textbf{Extended Data \cref{suppfigure:mimic_dataset_analysis_flowchart}}.
In both stages we use the AdamW optimizer with a learning rate of $1\times10^{-4}$ and a weight decay of 0.01. In addition gradient clipping (maximum norm = 1.0) and mixed precision training are employed for stability and efficiency. For complete training configurations please refer to \textbf{Extended Data Table~\ref{supptable:training_config}}.

\noindent
For various downstream tasks, ChexGen can be fine-tuned to enhance its generation capabilities (annotation to image). 
The training utilizes various spatial annotation datasets (see \textbf{Extended Data Table~\ref{supptable:dataset_stats}}) and follows the same optimization settings as the text to image training (two stage pretraining), but with spatial conditions incorporated into the diffusion objective. The model is trained at a resolution of 512$\times$512 with a batch size of 32 per GPU across 32 A100 GPUs.
Please refer to \textbf{Extended Data Table~\ref{supptable:training_config}} for complete training configurations.

\noindent

\heading{Quality Assessment of Generated Images}

\noindent
We conducted comprehensive quantitative evaluations on a held-out test set $\mathcal{D}_{\text{Test, PA}}$ of 3,500 posterior anterior (PA) view cases from the p19 subfolder of the MIMIC-CXR dataset. 
Each case includes a chest radiograph and its corresponding radiological report. 
For fair comparisons across different models, we used the impression sections from these radiological reports as generation prompts.
We evaluated both versions of our model against state-of-the-art medical image generation models RoentGen~\cite{chambon2022roentgen} and MINIM~\cite{wang2024self}. 
We employed three complementary metrics to assess different aspects of generation quality, including Fr\'{e}chet Inception Distance (FID), Structural Similarity Index (SSIM) and Pearson correlation.

\noindent
For fidelity assessment, we computed the FID between synthetic and real images using features extracted from two complementary models: a general purpose CLIP model~\cite{radford2021clip} that captures high level visual semantics with 768 dimensional features, and a domain specific CXR DenseNet121 classification model~\cite{Cohen2022xrv} (densenet121-res224-all) that extracts 512 dimensional medical relevant features.
Both feature vectors are obtained from the final average pooling layer of their networks. 
This evaluation strategy provides comprehensive insights into both general image quality and the preservation of domain specific medical characteristics, as CLIP assesses overall visual fidelity while the specialized CXR DenseNet121 model focuses on clinically relevant features.

\noindent
To evaluate diversity, we computed pairwise SSIM between synthetic images. For each test impression, we generate four images using different random seeds while keeping all other generation parameters fixed.
The set of generated images for each test impression is used to calculate pairwise SSIM values, which are then averaged to yield a case-level SSIM score.
Lower SSIM values indicate higher diversity, as they suggest less structural similarity between images generated from the same prompt, helping us assess the model's ability to capture the natural variability in medical imaging.

\noindent
For factual correctness evaluation, we employed a pretrained CXR-DenseNet121 model~\cite{Cohen2022xrv} (densenet121-res224-mimic) to predict 14 common chest pathologies on both real and synthetic images.
We computed the Pearson correlation coefficient between the prediction scores and the ground truths (dimension: $N \times 14$, where $N = 3500$) to assess how well the synthetic images preserve the pathological features present in real images. 
This metric is particularly important for medical applications as it directly measures the preservation of clinically relevant features.

\noindent
We utilized the t-SNE method to further visualize the distribution between synthetic and real images in feature space. First, we extract features using the pretrained CXR-DenseNet121 model~\cite{Cohen2022xrv} (densenet121-res224-mimic).
Then, these high-dimensional features are projected into a 2D space using t-SNE visualization. This analysis helps validate whether the synthetic images maintain similar feature distributions as real images, particularly in terms of pathological characteristics.

\noindent
To assess the clinical fidelity of generated images, we conducted a systematic radiological evaluation with two board certified radiologists (with 10 and 20 years of clinical experience, respectively). 
The evaluation compares our ChexGen model against state-of-the-art RoentGen using 150 randomly selected cases from the MIMIC-CXR test set. 
For each case, the radiologists are presented with the original impression report and the corresponding synthetic images from both models in a randomized and blinded manner. 
The assessment employs a 5 point Likert scale ranging from $-2$ to $+2$, where $-2$ indicates complete mismatch between the image and report, $-1$ indicates missing key pathological findings, $0$ indicates presence of major findings but missing details, $1$ indicates good alignment with most aspects of the report, and $2$ indicates perfect correspondence with all reported findings. 
Each image was evaluated independently, with attention given to the presence and accurate representation of pathological findings, anatomical structures, and their spatial relationships as described in the impression reports.

\heading{Training data augmentation experiments}

\noindent
To systematically evaluate the utility of synthetic images as training data supplements, we conducted extensive experiments across four key medical imaging tasks: disease classification, anatomical segmentation, lesion detection, and severity score regression. 
For each task, we generated synthetic images using our ChexGen model conditioned on task-specific data from the original training set (e.g., disease descriptions, segmentation masks, bounding boxes, or severity annotations). 
These synthetic images were then combined with the original training data at either a 1$\times$ or 2$\times$ scale, ensuring that the label distribution remained similar to that of the original dataset. 
All downstream models were trained from scratch using the mixed dataset (real + synthetic) and evaluated on the original test sets to assess the impact of synthetic data augmentation.
Representative examples of synthetic images for each task are shown in \textbf{Extended Data Figures~\ref{suppfigure:cond_gen_cls_example}, \ref{suppfigure:cond_gen_seg_example}, \ref{suppfigure:cond_gen_det_example_object}, and \ref{suppfigure:cond_gen_det_example_object_chex10}}. Detailed performance metrics across all tasks are reported in \textbf{Extended Data Tables~\ref{supptable:performance_classification}, \ref{supptable:performance_segmentation}, \ref{supptable:performance_detection}, and \ref{supptable:performance_regression}}. 
A summary of the datasets used can be found in \textbf{Extended Data Table~\ref{supptable:downstream_stats}}.

\hheading{Disease Classification}

\noindent
We assessed whether data generated by ChexGen can enhance the performance of downstream classification tasks.
We utilized DenseNet121~\cite{huang2017densely} as the backbone network for multi-label chest pathology classification. The network processes images resized to 256$\times$256 pixels and generates predictions for 14 pathology classes. 
We evaluated the model's performance on both internal and external test sets, using the state-of-the-art RoentGen as a baseline for comparison.
The classification model is trained on a dataset, \( \mathcal{D}_{\text{Train, PA}} \), consisting of 45,000 curated PA-view chest X-rays from MIMIC-CXR.
This dataset preserves the natural long-tail distribution of disease categories, ranging from common conditions such as Atelectasis (6,333 cases) and Cardiomegaly (5,678 cases) to rarer conditions like Pneumothorax (1,471 cases) and Enlarged Cardiomediastinum (829 cases).
The training set is augmented with synthetic images generated by ChexGen and RoentGen while maintaining the original class distribution.
The internal test set \( \mathcal{D}_{\text{Test, PA}} \) contains 3,500 cases from the MIMIC-CXR.
External evaluation is performed on the VinDr-CXR test set\cite{nguyen2022vindr}, which comprises 3,000 cases covering six overlapping disease categories with MIMIC-CXR: Atelectasis, Cardiomegaly, Consolidation, Pleural Effusion, Pneumonia, and Pneumothorax.
Performance is quantified using the macro-averaged Area Under the Receiver Operating Characteristic curve (macro-AUROC).
Training is conducted for 40 epochs using an AdamW optimizer with a batch size of 128. To mitigate class imbalance, we employ weighted sampling to ensure that all pathology classes are equally represented during training.

\hheading{Anatomical Segmentation}

\noindent
We assessed whether ChexGen-generated data can enhance the performance of downstream anatomical segmentation tasks. 
We employed a standard UNet\cite{ronneberger2015u} with a ResNet34\cite{he2016deep}, which is widely used for medical image segmentation, and evaluated the model's performance on both internal and external test sets.
We utilized the SIIM-ACR-PTX pneumothorax segmentation dataset, which contains 12,047 chest radiographs with expert-annotated pixel-level masks. 
The dataset is divided into 8,433 training cases and 2,407 test cases.
We augmented the training set with synthetic images generated by ChexGen while maintaining the original distribution of positive and negative cases. 
Follow the official guidance, model performance was assessed using two complementary metrics: the positive Dice coefficient (computed only on cases with pneumothorax) and the negative Dice coefficient (computed on cases without pneumothorax).
To assess zero-shot generalization capabilities, we conducted an external test on the Candid-PTX dataset, which comprises 19,237 cases.
Additionally, we used the same UNet model to segment different anatomical structures, such as ribs, using the VinDr-RibCXR dataset. This dataset contains 245 cases (196 for training and 49 for testing) with detailed rib cage annotations.
Performance on this anatomically distinct task was measured using the standard Dice similarity coefficient.
Throughout all segmentation experiments, the model processed input images at a 512$\times$512 resolution with a batch size of 32. The model was trained for 40 epochs using an AdamW optimizer with a learning rate of 0.001.

\hheading{Abnormality Localization}

\noindent
We assessed whether ChexGen-generated data can enhance the performance of downstream abnormality localization tasks.
For this purpose, we employed a Faster R-CNN model\cite{ren2015faster} with a ResNet-50 backbone\cite{he2016deep} and Feature Pyramid Network (FPN)\cite{lin2017feature}. 
We developed and tested the model on the Object-CXR and ChesX-Det datasets\cite{objectcxr,liu2020chestxdet10}. 
Object-CXR contains 9,000 chest radiographs (8,000 for training and 1,000 for testing) with bounding box annotations for foreign object detection. 
ChesX-Det comprises 5,213 chest radiographs (4,720 for training and 493 for testing), we utilized annotations for seven common thoracic conditions: No Finding, Atelectasis, Consolidation, Effusion, Emphysema, Fibrosis, and Pneumothorax.
The model is trained with a batch size of 8 over 70k training iterations using an input size of 1000$\times$1000 pixels. The base learning rate is set to 0.001.
Following standard practices in medical object detection, we evaluate model performance using mean Average Precision (mAP50) and mean Average Recall (mAR50) at an Intersection over Union (IoU) threshold of 0.5 between the predicted bounding boxes and the ground truth annotations.

\hheading{Severity Score Regression}

\noindent
We assessed whether ChexGen-generated data can enhance the performance of downstream regression tasks.
For this purpose, we employed a ResNet50 backbone followed by a regression head. 
We developed and tested the model on the Stonybrook Radiographic Assessment of Lung Opacity (RALO) dataset\cite{cohen2021radiographic}, which comprises 2,373 chest radiographs (1,898 for training and 475 for testing) with expert-annotated opacity severity scores. 
For each case, two expert-provided Lung Opacity scores are available, and we use their mean as the regression target. The model processes input images at a resolution of 224$\times$224 with a batch size of 32. 
Training is conducted for 50 epochs using the AdamW optimizer with a learning rate of 0.001. Performance is evaluated using Mean Absolute Error (MAE), Mean Squared Error (MSE), and Pearson correlation coefficient.

\heading{Supervised pretraining experiments}

\noindent
While synthetic data shows great potential in medical imaging, directly combining synthetic data with limited real data for joint training may lead to performance degradation due to distribution shifts, especially in scenarios with extremely limited real data.
To address this issue, we designed a two-stage training strategy: supervised pretraining on synthetic data followed by fine-tuning on real data.
A summary of the datasets used can be found in \textbf{Extended Data Table~\ref{supptable:downstream_stats}}.

\noindent
In the pretraining stage, we utilized a subset \( \mathcal{D}_{\text{Train, PA}} \) of the MIMIC-CXR dataset, which comprises 45,000 chest radiographs with corresponding impression texts. 
Using these texts as prompts, we generated synthetic images at multiple scales (1×, 2×, 3×, and 5× the original dataset size) while preserving the original pathology distribution. 
Each synthetic image inherits the disease labels from its corresponding source case, thus creating paired training data for supervised pretraining.
For the pretraining phase, we employed a DenseNet121 backbone initialized with ImageNet weights. 
All images are resized to 256×256 pixels and augmented using standard techniques, including random horizontal flips, rotations (\(\pm10^\circ\)), and intensity adjustments. 
We trained the model for 100 epochs using the AdamW optimizer with \(\beta = (0.9, 0.999)\) and a learning rate of \(1\times10^{-4}\) following a cosine decay schedule. 
Training is conducted on a single NVIDIA A100 GPU with a batch size of 128.
After pretraining, we evaluated the model's performance on MIMIC-CXR by fine-tuning it with varying percentages of the real training data (1\%, 10\%, 30\%, 50\%, 70\% and 100\%). 
For fine-tuning, we use a reduced learning rate of \(1\times10^{-5}\) and a batch size of 32, training the model for 40 epochs.
Model performance was evaluated on the held-out test set \( \mathcal{D}_{\text{Test, PA}} \) consisting of 3,500 cases, and the detailed performance metrics are reported in \textbf{Extended Data Figure~\ref{suppfigure:pretrain_mimic}}.

\noindent
To further validate generalization capability, we perform a cross-domain evaluation by transferring our pretrained models to both the MedFMC-ChestDR\cite{wu2023radfm} few-shot learning benchmark and the VinDr-CXR dataset\cite{nguyen2022vindr}. 
The MedFMC-ChestDR dataset comprises 4,848 frontal chest radiographs collected from two regional hospitals in China, with expert annotations for 19 common thoracic diseases. 
Following the official MedFMC evaluation protocol, we evaluate the models under 1-shot, 5-shot, and 10-shot scenarios, where each "shot" represents the number of labeled examples per class available for fine-tuning, and model performance is measured on an official test set of 800 images. 
Similarly, the VinDr-CXR dataset contains 15,000 images (12000 training cases, 3000 testing cases) spanning 15 common thoracic diseases categories; for this dataset, we fine-tune the models using different proportions of the training data (1\%, 10\%, 30\%, 50\%, 70\%, and 100\%) and assess performance on the corresponding test set.
This evaluation setting helps assess whether the representations learned during synthetic data pretraining can effectively transfer to new domains with minimal real data supervision.
Please refer to \textbf{Extended Data Figure~\ref{suppfigure:pretrain_medfmc} and \cref{suppfigure:pretrain_vindr}} for the detailed performance metrics.

\heading{Model bias detection and mitigation experiments}

\noindent
Following the methodology\cite{glocker2023algorithmic}, we obtained demographic information (age and race) for each study in the MIMIC-CXR dataset by linking it to the MIMIC-IV database.
We simplify the demographic categorization by focusing on two gender groups (Male and Female) and three racial groups (White and Black) and three age groups (19–40, 41–65, and over 66). 
After applying these demographic filters, our refined dataset comprises 31,475 training cases and 2,415 test cases, drawn from the original MIMIC-CXR training set \( \mathcal{D}_{\text{Train, PA}} \) and test set \( \mathcal{D}_{\text{Test, PA}} \).
Using ChexGen, we synthesize a matched test set that preserves each real case's demographic attributes and clinical findings, enabling direct comparisons between real and synthetic evaluations. 
To systematically assess model performance, we analyze it across individual demographic factors (e.g., race and age) as well as their intersections (e.g., performance for White individuals aged 66+ and Black individuals aged 19-40).

\noindent
To demonstrate that synthetic data can reliably be used to measure model performance, we first trained a DenseNet121 model (initialized with ImageNet weights) on the original training set and then evaluated it on both real and synthetic test sets.
For each demographic group, we computed the weighted F1-score across 14 disease categories. 
Finally, we calculated Pearson correlation coefficients between F1-scores from real and synthetic data across different demographic groups to assess the reliability of the synthetic evaluation.

\noindent
After validating synthetic data's efficacy for bias detection, we explored its potential for bias mitigation by addressing the demographic imbalance in the MIMIC-CXR dataset. 
Using ChexGen, we generated additional training samples specifically for underrepresented demographic subgroups, creating chest X-rays conditioned on both clinical findings and demographic attributes until achieving demographic parity with majority groups. 
A new de-biased model (DenseNet121, initialized with ImageNet weights) was then trained on this demographically balanced dataset and evaluated on the same real test set as the original biased model. 
To quantitatively assess fairness improvements, we compared two key metrics across all demographic subgroups: the weighted F1-score of the worst-performing group and the performance gap between the best and worst performing groups. 

\heading{Statistical analysis}

\noindent
All experiments were conducted over 5 independent experiments, and error bars represent the standard deviation across these runs. For paired model comparisons, we used a two-sided Wilcoxon signed-rank test to determine statistical significance. Additionally, we employed the Pearson correlation coefficient and its associated two-sided p-value (computed using scikit-learn's default method) to assess correlations between continuous variables. A p-value below 0.05 was considered statistically significant.

\heading{Computing Hardware and Software}

\noindent
We implemented all experiments using Python (version 3.8.13) with PyTorch (version 2.2.0, CUDA 12.1) as the primary deep learning framework. The diffusion model training was conducted on 32 80GB NVIDIA A100 GPUs using DistributedDataParallel (DDP), while downstream evaluations were performed on single NVIDIA A100 GPUs.
For the diffusion model backbone, we utilized the open-source diffusers library (version 0.26.3) from Hugging Face, which provides implementations of common diffusion model architectures and training utilities. The DiT architecture was adapted from the official DiT report's XL/2 configuration, while the variational autoencoder (VAE) was adopted from Stable Diffusion's pretrained VAE. For condition embedding, we used the transformers library (version 4.38.1) to process text inputs, employing the official T5 model for text encoding. To optimize memory usage and training efficiency, we incorporated xformers (version 0.0.24) for efficient attention computation.
Image processing and data augmentation were handled using albumentations (version 1.4.2) and torchvision (version 0.17.0). Medical image specific processing, including DICOM handling and window/level adjustments, was implemented using pydicom (version 2.4.4), SimpleITK (version 2.3.1), and OpenCV (version 4.9.0). For evaluation metrics, we employed torchmetrics (version 0.11.4) for standard classification metrics, MONAI (version 1.3.0) for medical image specific metrics such as Dice scores, and scikit-learn (version 1.4.1) for general machine learning metrics.
Data management and analysis were conducted using NumPy (version 1.26.4) and Pandas (version 2.2.1). Visualization of results and model outputs was performed using Matplotlib (version 3.8.3) and seaborn (version 0.13.2). For experiment tracking and logging, we used Weights \& Biases (version 0.15.0) and TensorBoard (version 2.16.2).

\end{spacing}

%%%%%%%%%%%%%%%%%
%%% Main Figures
%%%%%%%%%%%%%%%%%
\clearpage
\renewcommand{\figurename}{Extended Data Figure}

% \begin{figure*}
%     \centering
%     \includegraphics[width=1.00\textwidth]{resources/data/supplements/longtail_medfmc.png}
%     \caption{
%     \textbf{Performance drop on long-tail disease categories.} 
%     Comprehensive performance comparison across all 19 disease categories in MedFMC dataset, demonstrating the cross-domain generalization capability of our approach. Using the model pretrained on MIMIC-CXR synthetic data, we evaluate its transferability to a completely different dataset (MedFMC). Compared to ImageNet pretraining baseline, our ChexGen pretraining shows consistent improvements across different few-shot learning scenarios (1-shot, 5-shot, and 10-shot). The results demonstrate that ChexGen pretraining not only significantly enhances the model's ability to learn from limited data but also successfully transfers knowledge across different medical datasets. \textcolor{red}{[is the text paired with the figure? i need a b c]}
%     }
%     \label{suppfigure:drop_medfmc}
% \end{figure*}

\clearpage    

\setcounter{table}{0}
\renewcommand{\tablename}{Extended Data Table}

% chexgen training

\begin{table}[t]
    \centering
    \renewcommand{\arraystretch}{1.15}
    \resizebox{\textwidth}{!}{
    \begin{tabular}{lcccccc}
        \toprule[1.5pt]
        \textbf{Dataset} (\( \mathcal{D} \)) & \textbf{\# Cases} & \textbf{Year} & \textbf{Label Classes} & \textbf{Label Type} & \textbf{Label Source} & \textbf{Region} \\
        \midrule[1pt]
        \multicolumn{7}{l}{\textbf{Stage I: Vision-Language Pretraining}} \\[0.5ex]
        NIH ChestX-ray14\cite{wang2017chestxray14} (\( \mathcal{D}_{\text{NIH}} \)) & 112,120 & 2017 & 14 diseases & Classification label & Human & USA \\
        Stanford Chexpert\cite{irvin2019chexpert} (\( \mathcal{D}_{\text{CheXpert}} \)) & 223,414 & 2019 & 14 diseases & Classification label & Human & USA \\
        PadChest\cite{bustos2020padchest} (\( \mathcal{D}_{\text{PadChest}} \)) & 166,737 & 2019 & 193 radiological findings & Classification label & Human & Spain \\
        MIMIC-CXR\cite{johnson2019mimiccxr} (\( \mathcal{D}_{\text{Train, Multi}} \)) & 332,524 & 2019 & 14 diseases & Classification label & Human & USA \\
        BIMCV-Covid\cite{vaya2020bimcv} (\( \mathcal{D}_{\text{BIMCV}} \)) & 45,633 & 2020 & 188 radiological findings & Classification label & Human & Spain \\
        VinDr-PCXR\cite{pham2022vindr} (\( \mathcal{D}_{\text{VinDr}} \)) & 7,728 & 2022 & 15 diseases & Classification label & Human & Vietnam \\
        BRAX\cite{reis2022brax} (\( \mathcal{D}_{\text{BRAX}} \)) & 40,967 & 2022 & 14 diseases & Classification label & Human & Brazil \\
        RANZCR Clip\cite{tang2021clip} (\( \mathcal{D}_{\text{RANZCR}} \)) & 30,083 & 2021 & 11 device placements & Classification label & Human & Australia \\[0.5ex] \hline
        OpenChest (\( \mathcal{D}_{\text{OpenChest}} \))& 959,206 & 2024 & Generic Descriptions & Text & AI (GPT4) & Multi-national \\[0.3ex]
        \midrule
        \multicolumn{7}{l}{\textbf{Stage II: Vision-Language Alignment}} \\[0.5ex]
        MIMIC-CXR\cite{johnson2019mimiccxr} (\( \mathcal{D}_{\text{Train, PA}} \)) & 45,000 & 2019 & Clinical Report & Text & Human & USA \\[0.5ex] 
        \bottomrule[1.5pt]
        \multicolumn{7}{l}{\footnotesize \textasteriskcentered ~Stage II MIMIC-CXR data is curated from the full dataset by selecting images with detailed reports with impression token length < 120} \\
    \end{tabular}
    }
    \caption{
    \textbf{Dataset statistics for text-to-image training pipeline.} Our training process consists of two stages: \textit{Stage I (Vision-Language Pretraining)} utilizes the OpenChest dataset (approximately 960K image-text pairs) for foundation pretraining, where GPT-4 converts medical metadata from multiple public datasets into standardized radiological descriptions. \textit{Stage II (Vision-Language Alignment)} employs clinical reports from the MIMIC-CXR dataset (45K pairs) written by professional radiologists for further alignment. For each dataset, we provide the number of cases, release year, label classes and type, label source, and geographical region. 
    }
    \label{supptable:dataset_stats}
\end{table}

\clearpage

\begin{table}[t]
    \centering
    \renewcommand{\arraystretch}{1.15}
    \resizebox{\textwidth}{!}{
    \begin{tabular}{lcccccc}
        \toprule[1.5pt]
        \textbf{Dataset} (\( \mathcal{D} \)) & \textbf{\# Cases} & \textbf{Year} & \textbf{Label Classes} & \textbf{Label Type} & \textbf{Label Source} & \textbf{Region} \\
        \midrule[1pt]
        \multicolumn{7}{l}{\textbf{Extending to Annotation-to-Image Generation}} \\[0.5ex]
        SIIM-ACR-PTX\cite{siim} (\( \mathcal{D}_{\text{SIIM}} \)) & 10,675 & 2019 & Pneumothorax & Segmentation mask & Human & USA \\
        VinDr-RibCXR\cite{nguyen2021vindr} (\( \mathcal{D}_{\text{VinDr-Rib}} \)) & 196 & 2021 & Rib & Segmentation mask & Human & Vietnam \\
        Object-CXR\cite{kufel2023chest} (\( \mathcal{D}_{\text{Object}} \)) & 3968 & 2023 & Foreign objects & Bounding box & Human & China \\
        ChestX-Det\cite{liu2020chestxdet10} (\( \mathcal{D}_{\text{ChesXDet}} \)) & 4720 & 2020 & 10 diseases & Bounding box & Human & USA \\
        RALO\cite{cohen2021radiographic} (\( \mathcal{D}_{\text{RALO}} \)) & 1898 & 2022 & Lung opacity & Assessment score & Human & USA \\[0.5ex]
        \bottomrule[1.5pt]
    \end{tabular}
    }
    \caption{
    \textbf{Dataset statistics for downstream annotation-to-image finetuning.} After the two-stage pretraining, we extend ChexGen to support annotation-guided image generation by incorporating datasets with diverse types of spatial annotations. These include segmentation masks for specific anatomical structures (e.g., pneumothorax and ribs) and bounding boxes for disease localization and foreign objects. For each dataset, we report the number of cases, release year, label classes and type, label source, and geographical region. 
    }
    \label{supptable:dataset_stats_2}
\end{table}

\clearpage
\begin{table}[t]
    \centering
    \resizebox{\textwidth}{!}{
    \begin{tabular}{lcccccc}
        \toprule[1.5pt]
        \textbf{Dataset} & \textbf{\# Cases (Train/Test)} & \textbf{Year} & \textbf{Label Classes} & \textbf{Label Type} & \textbf{Task Description} & \textbf{Metric} \\
        \midrule[1pt]
        \multicolumn{7}{l}{\textbf{Data Augmentation}} \\[0.5ex]
        MIMIC-CXR\cite{johnson2019mimiccxr} & 45,000 / 3,500 & 2019 & 14 diseases & Classification & Multi-label disease classification & AUROC \\
        VinDr-CXR\cite{pham2022vindr} & -- / 3,000 & 2022 & 6 diseases\textsuperscript{\textdagger} & Classification & External validation on disease classification & AUROC \\
        SIIM-ACR-PTX\cite{siim} & 10,675 / 1,372 & 2019 & Pneumothorax & Segmentation & Pneumothorax segmentation & Dice \\
        Candix-PTX\cite{nguyen2021vindr} & -- / 19,237 & 2021 & Pneumothorax & Segmentation & External validation on pneumothorax segmentation & Dice \\
        VinDr-RibCXR\cite{nguyen2021vindr} & 196 / 49 & 2021 & Rib & Segmentation & Rib segmentation & Dice \\
        Object-CXR\cite{kufel2023chest} & 8,000 / 1,000 & 2023 & Foreign objects & Detection & Foreign object detection & mAP50, mAR50 \\
        ChestX-Det\cite{liu2020chestxdet10} & 4,720 / 475 & 2020 & 10 diseases & Detection & Chest disease localization & mAP50, mAR50 \\
        RALO\cite{cohen2021radiographic} & 1,898 / 475 & 2022 & Lung opacity & Assessment score & Severity scoring & MAE, MSE, Pearson's r \\[0.5ex]
        \midrule
        \multicolumn{7}{l}{\textbf{Supervised Pretraining}} \\[0.5ex]
        MIMIC-CXR\cite{johnson2019mimiccxr} & \textsuperscript{\textasteriskcentered}/ 3,500 & 2019 & 14 diseases & Classification & Limited-data disease classification & AUROC \\
        MedFMC-ChestDR\cite{wang2023real} & \textsuperscript{\textdaggerdbl}/ 400 & 2023 & 19 diseases & Classification & Few-shot medical classification & AUROC \\
        VinDr-CXR\cite{nguyen2022vindr} & 
        \textsuperscript{\textbullet}/ 3,000 & 2022 & 15 diseases & Classification & Limited-data disease classification & AUROC 
        
        \\[0.5ex]
        \midrule
        \multicolumn{7}{l}{\textbf{Model Evaluation}} \\[0.5ex] 
        MIMIC-CXR\cite{johnson2019mimiccxr} & 31,475 / 2,415 & 2019 & 14 diseases & Classification &Multi-label disease classification  & AUROC \\
        \bottomrule[1.5pt]
        \multicolumn{7}{l}{\footnotesize \textdagger ~Overlapping diseases with MIMIC-CXR: Atelectasis, Cardiomegaly, Consolidation, Lung Opacity, Pleural Effusion, Pneumothorax} \\
        \multicolumn{7}{l}{\footnotesize \textasteriskcentered ~MIMIC-CXR varying training set sizes: 1\% (450), 10\% (4500), 30\% (13500), 50\% (22500), 50\% (31499), 100\% (45,000)} \\ 
        \multicolumn{7}{l}{\footnotesize \textdaggerdbl ~MedFMC few-shot training settings: 1-shot (19), 5-shot (95), 10-shot (190)} \\
        \multicolumn{7}{l}{\footnotesize \textbullet   ~Vindr-CXR varying training set sizes: 1\% (122), 10\% (1,215), 30\% (3,615), 50\% (5,985), 70\% (8,385) 100\% (12,000)} 
    \end{tabular}
    }
    \caption{
    \textbf{Dataset statistics for downstream applications.} We evaluated ChexGen's effectiveness across three key applications: (1) Data Augmentation, using synthetic data to enhance performance on various medical tasks including classification, segmentation, detection, and severity scoring; (2) Supervised Pretraining, investigating the model's data efficiency through limited-data and few-shot learning scenarios on MIMIC-CXR and MedFMC-Chest datasets; and (3) Model Evaluation, assessing demographic fairness on the MIMIC-CXR dataset. For each dataset, we provide train/test splits, task descriptions, and evaluation metrics. 
    }
    \label{supptable:downstream_stats}
\end{table}
\clearpage

\begin{table}[t]
    \centering
    \resizebox{\textwidth}{!}{
    \begin{tabular}{lcc|c}
    \toprule
    Configuration & Stage I: Vision-Language Pretraining & Stage II: Vision-Language Alignment & Annotation-to-Image Generation \\
    \midrule
    \multicolumn{4}{l}{\textit{Model Components}} \\
    VAE Init. & SD-v1.5* & from Stage 1* & from Stage 2* \\
    Text Encoder Init. & T5* & from Stage 1* & from Stage 2* \\
    Diffusion Transformer & DiT-XL/2 & from Stage 1 & from Stage 2* \\
    ControlNet & - & - & Initialized \& Tuned \\
    \midrule
    \multicolumn{4}{l}{\textit{Data Config}} \\
    Image Resolution & $256^2$ & $512^2$ & $512^2$ \\
    Text Length & 120 & 120 & 120 \\
    \midrule
    \multicolumn{4}{l}{\textit{Training Config}} \\
    Number of GPUs & 32 A100 & 32 A100 & 32 A100 \\
    Batch Size & 192 & 64 & 32 \\
    Training Epochs & 800 & 300 & 200 \\
    Optimizer & AdamW & AdamW & AdamW \\
    Learning Rate & 1e-4 & 1e-4 & 1e-4 \\
    Weight Decay & 0.01 & 0.01 & 0.01 \\
    Gradient Clip & 1.0 & 1.0 & 1.0 \\
    Mixed Precision & bf16 & bf16 & bf16 \\
    \bottomrule
    \end{tabular}
    }
    \caption{\textbf{Training hyperparameters across different stages}. Our training pipeline consists of three stages with distinct configurations: Stage I (Vision-Language Pretraining) is conducted at 256$\times$256 resolution; Stage II (Vision-Language Alignment) inherits and fine-tunes components from Stage I at 512$\times$512 resolution; and the Annotation-to-Image Generation stage introduces and trains the ControlNet adapter while keeping other components frozen. All stages employ consistent optimization settings (AdamW, learning rate 1e-4, weight decay 0.01) and utilize 32 A100 GPUs. * indicates frozen parameters during training. }
    \label{supptable:training_config}
\end{table}

\clearpage
\begin{table}[t]
    \centering
    \renewcommand{\arraystretch}{1.2}
    \setlength{\tabcolsep}{10pt}
    \begin{tabular}{|p{0.9\textwidth}|}
        \hline
        \textbf{Patient Meta Information:} \\
        Age: 65 years; Gender: Male; Race: White; X-ray View: Frontal \\
        Diagnosis: Pneumonia, Pleural Effusion, Airspace Opacity; ...... \\
        \hline
        \textbf{Curation Query:} \\
        Please generate a comprehensive radiology report based on the provided patient information. Your report should follow the standard radiology report format with the following sections:
            1. Patient Information: Include all relevant patient demographic information (age, gender, race).
            2. Technique: Describe the imaging modality and protocol used.
            3. Findings: Present the findings as a cohesive paragraph, not a list; Include all abnormalities exactly as provided in the metadata (e.g., "Pneumonia"); For each abnormality, include its typical appearance using standard terminology; Do not infer additional information such as severity, distribution, or location unless explicitly mentioned in metadata; Avoid any speculative interpretation or clinical correlation beyond what is stated.
            4. Impression: Provide a brief summary restating the key abnormalities in metadata.
            5. Ensure information is accurately represented without speculation beyond the provided data. \\ \hline
        \textbf{GPT-4 Response:}  
        \begin{itemize}
            \setlength{\itemsep}{-1ex}
            \item Patient Information: A white male patient aged 65 years.
            \item Technique: Frontal chest radiograph was performed.
            \item Findings: The frontal chest radiograph demonstrates airspace opacity, indicative of alveolar filling processes, which appears as regions of increased pulmonary attenuation. Findings are consistent with pneumonia, typically manifested as focal or multifocal airspace opacities. %Additionally, there is evidence of pleural effusion, seen as fluid accumulation with potential blunting of the costophrenic angles.
            \item Impression: Findings are consistent with pneumonia, pleural effusion, and airspace opacity.
        \end{itemize} \\ \hline
        \textbf{Quality Control Query:} \\
        Please review the generated report against the patient meta information and provide a corrected version that: 
        1. Identifies and corrects any discrepancies;
        2. Removes redundant information;
        3. Supplements missing details;
        4. Ensures medical accuracy and completeness;
        5. Please output the revised report directly. \\ \hline
        \textbf{Qwen-2.5 Response:} 
        \begin{itemize}
            \setlength{\itemsep}{-1ex}
            \item Patient Information: A white male patient aged 65 years.
            \item Technique: Frontal chest radiograph was performed.
            \item Findings: The frontal chest radiograph demonstrates airspace opacity, indicative of alveolar filling processes, which appears as regions of increased pulmonary attenuation. Findings are consistent with pneumonia, typically manifested as focal or multifocal airspace opacities. \colorbox{gray!10}{\parbox{\dimexpr\linewidth-2\fboxsep}{
            Additionally, there is evidence of pleural effusion, seen as fluid accumulation with potential blunting of the costophrenic angles.
            }}
            \item Impression: Findings are consistent with pneumonia, pleural effusion, and airspace opacity.
        \end{itemize} \\ \hline
    \end{tabular}
    \caption{\textbf{Example of OpenChest dataset curation using LLMs}. This example illustrates our data curation process: (1) Patient meta-information containing structured clinical data such as demographics, medical history, symptoms, and test records; (2) A detailed prompt template that guides GPT-4 to generate standardized medical reports by specifying key requirements for content inclusion, accuracy, and format; (3) The generated response that synthesizes the provided information into a professional medical report; and (4) A quality control step using Qwen-2.5 to review and enhance the initial GPT-4 generated report, ensuring accuracy, completeness, and clinical relevance by identifying discrepancies, removing redundancies, and supplementing missing information. }
    \label{supptable:openchest_curation}
\end{table}
% data augmentation, mimic-cxr
\begin{table}[t]
\centering
\resizebox{\textwidth}{!}{%
\begin{tabular}{lccccc}
\toprule
Category & Real & \makecell{Real+1$\times$Syn. \\ (ChexGen)} & \makecell{Real+2$\times$Syn. \\ (ChexGen)} & \makecell{Real+1$\times$Syn. \\ (RoentGen)} & \makecell{Real+2$\times$Syn. \\ (RoentGen)} \\
\midrule
\multicolumn{6}{c}{Internal Validation (MIMIC-CXR)} \\
\midrule
No Finding          & $84.93 \pm 0.16$ & $85.41 \pm 0.14$ & $\bm{85.49 \pm 0.25}$ & $85.35 \pm 0.11$ & $85.36 \pm 0.08$ \\
Enlarged Cardiome.  & $52.86 \pm 0.31$ & $60.97 \pm 1.92$ & $\bm{62.27 \pm 1.00}$ & $56.18 \pm 2.28$ & $54.16 \pm 0.72$ \\
Cardiomegaly        & $85.40 \pm 0.45$ & $\bm{86.06 \pm 0.39}$ & $86.05 \pm 0.50$ & $84.14 \pm 0.23$ & $85.80 \pm 0.18$ \\
Airspace Opacity    & $73.08 \pm 0.31$ & $\bm{74.15 \pm 0.10}$ & $74.05 \pm 0.04$ & $73.03 \pm 0.16$ & $73.85 \pm 0.19$ \\
Lung Lesion         & $67.36 \pm 0.61$ & $70.27 \pm 1.37$ & $\bm{70.77 \pm 0.46}$ & $67.58 \pm 1.15$ & $67.23 \pm 0.26$ \\
Edema               & $92.45 \pm 0.28$ & $\bm{92.84 \pm 0.28}$ & $92.76 \pm 0.22$ & $92.70 \pm 0.03$ & $92.45 \pm 0.24$ \\
Consolidation       & $76.09 \pm 0.53$ & $77.84 \pm 0.89$ & $\bm{77.86 \pm 0.65}$ & $76.77 \pm 0.22$ & $76.83 \pm 0.62$ \\
Pneumonia           & $71.18 \pm 0.49$ & $72.02 \pm 0.44$ & $\bm{72.25 \pm 0.33}$ & $71.97 \pm 0.40$ & $72.03 \pm 0.48$ \\
Atelectasis         & $80.39 \pm 0.20$ & $81.19 \pm 0.29$ & $\bm{81.56 \pm 0.33}$ & $81.20 \pm 0.27$ & $80.87 \pm 0.16$ \\
Pneumothorax        & $90.62 \pm 1.01$ & $91.33 \pm 0.47$ & $\bm{91.85 \pm 0.45}$ & $91.28 \pm 0.12$ & $91.10 \pm 0.50$ \\
Pleural Effusion    & $92.91 \pm 0.18$ & $93.24 \pm 0.18$ & $\bm{93.46 \pm 0.14}$ & $92.20 \pm 0.08$ & $92.97 \pm 0.16$ \\
Pleural Other       & $76.48 \pm 1.62$ & $79.62 \pm 1.08$ & $\bm{79.71 \pm 1.12}$ & $77.73 \pm 1.11$ & $77.54 \pm 1.15$ \\
Fracture            & $68.35 \pm 0.93$ & $69.67 \pm 1.56$ & $\bm{71.46 \pm 1.34}$ & $69.62 \pm 0.36$ & $69.31 \pm 0.77$ \\
Support Devices     & $80.54 \pm 0.68$ & $81.40 \pm 0.53$ & $\bm{81.85 \pm 0.40}$ & $80.14 \pm 0.23$ & $80.05 \pm 0.43$ \\
\midrule
Average             & $78.05 \pm 0.28$ & $79.72 \pm 0.02$ & $\bm{80.10 \pm 0.14}$ & $78.56 \pm 0.27$ & $78.54 \pm 0.16$ \\
\midrule
\multicolumn{6}{c}{External Test (VinDr-CXR)} \\
\midrule
Pleural Effusion    & $79.29 \pm 0.32$ & $\bm{80.49 \pm 0.21}$ & $80.31 \pm 0.15$ & $79.25 \pm 0.20$ & $79.40 \pm 0.38$ \\
Lung Opacity        & $62.33 \pm 1.25$ & $64.23 \pm 1.12$ & $\bm{65.04 \pm 1.69}$ & $61.23 \pm 1.12$ & $62.30 \pm 1.41$ \\
Atelectasis         & $46.75 \pm 2.57$ & $47.67 \pm 0.45$ & $\bm{50.56 \pm 0.57}$ & $46.85 \pm 1.23$ & $47.12 \pm 2.52$ \\
Cardiomegaly        & $76.96 \pm 1.82$ & $\bm{80.17 \pm 1.65}$ & $79.36 \pm 2.39$ & $77.25 \pm 2.10$ & $77.81 \pm 3.61$ \\
Pneumothorax        & $81.67 \pm 2.45$ & $\bm{84.03 \pm 0.95}$ & $83.37 \pm 0.71$ & $82.07 \pm 0.31$ & $82.10 \pm 0.20$ \\
Consolidation       & $77.67 \pm 1.28$ & $80.42 \pm 0.82$ & $\bm{81.69 \pm 0.97}$ & $77.95 \pm 0.54$ & $78.23 \pm 0.67$ \\
\midrule
Average             & $70.78 \pm 0.77$ & $72.84 \pm 0.45$ & $\bm{73.39 \pm 0.42}$ & $70.77 \pm 0.45$ & $71.16 \pm 0.71$ \\
\bottomrule
\end{tabular}%
}
\caption{\textbf{Evaluation of synthetic data for training data augmentation in disease classification.} Internal validation on MIMIC-CXR using 45,000 training images and a held-out test set of 3,500 images for 14 chest diseases. Zero-shot transfer performance on VinDr-CXR, where the model trained on MIMIC-CXR with varying ratios of synthetic data is directly evaluated on 3,000 test images without fine-tuning for 6 overlapping diseases that appear in both datasets. "Real + n$\times$Syn." indicates training with 1 part real data and n parts synthetic data. All results are reported as mean $\pm$ standard deviation over 5 independent runs, with the best performance in each row highlighted in bold. }
\label{supptable:performance_classification}
\end{table}
\clearpage

\begin{table}[t]
  \centering
  \begin{adjustbox}{max width=\textwidth}
  \begin{tabular}{l cc cc cc}
  \toprule
   & \multicolumn{2}{c}{Real} 
   & \multicolumn{2}{c}{Real+1$\times$Syn.} 
   & \multicolumn{2}{c}{Real+2$\times$Syn.} \\
  \cmidrule(lr){2-3} \cmidrule(lr){4-5} \cmidrule(lr){6-7}
  Category 
   & mAP50 & mAR50 
   & mAP50 & mAR50 
   & mAP50 & mAR50 \\
  \midrule
  \multicolumn{7}{c}{ChesX-Det} \\
  \midrule
  No Finding 
    & 69.38$\pm$0.27 & 92.70$\pm$1.37 
    & 72.88$\pm$0.38 & 97.99$\pm$0.69 
    & \textbf{74.61$\pm$0.68} & \textbf{98.14$\pm$0.58} \\
  Atelectasis 
    & 27.73$\pm$1.32 & 49.68$\pm$0.59 
    & \textbf{35.21$\pm$2.56} & \textbf{51.63$\pm$1.13} 
    & 33.19$\pm$1.04 & 65.16$\pm$0.64 \\
  Consolidation 
    & 57.95$\pm$1.64 & 72.31$\pm$0.71 
    & 58.09$\pm$0.90 & 73.47$\pm$1.42 
    & \textbf{60.75$\pm$0.51} & \textbf{85.13$\pm$1.19} \\
  Effusion 
    & 51.90$\pm$1.61 & 67.78$\pm$1.36 
    & 52.63$\pm$2.16 & 73.03$\pm$1.01 
    & \textbf{53.73$\pm$0.52} & \textbf{87.12$\pm$1.94} \\
  Emphysema 
    & 64.21$\pm$1.05 & 75.32$\pm$1.12 
    & 64.51$\pm$1.52 & 80.81$\pm$1.74 
    & \textbf{65.22$\pm$0.56} & \textbf{84.73$\pm$0.21} \\
  Fibrosis 
    & \textbf{40.95$\pm$1.62} & \textbf{58.21$\pm$0.97} 
    & 36.89$\pm$0.58 & 57.78$\pm$3.93 
    & 36.80$\pm$1.30 & 67.47$\pm$1.47 \\
  Pneumothorax 
    & 22.86$\pm$2.61 & 26.27$\pm$2.85 
    & 22.85$\pm$1.40 & 36.51$\pm$1.37 
    & \textbf{25.43$\pm$3.25} & \textbf{46.71$\pm$4.18} \\
  \midrule
  Overall 
    & 47.85$\pm$0.82 & 63.18$\pm$0.17 
    & 49.01$\pm$0.02 & 67.32$\pm$0.68 
    & \textbf{49.96$\pm$0.69} & \textbf{76.35$\pm$0.45} \\
  \midrule
  \multicolumn{7}{c}{Object-CXR} \\
  \midrule
  Foreign Object 
    & 61.02$\pm$0.31 & 90.60$\pm$0.15 
    & 62.70$\pm$0.02 & 93.69$\pm$0.28 
    & \textbf{63.59$\pm$0.70} & \textbf{94.38$\pm$0.28} \\
  \bottomrule
  \end{tabular}
  \end{adjustbox}
  \caption{\textbf{Evaluation of synthetic data for training data augmentation in abnormality localization.} Detection models were trained with real data alone and with synthetic data augmentation (1$\times$ and 2$\times$) on the ChesX-Det dataset for thoracic pathology detection and Object-CXR dataset for foreign object detection. Performance is measured using Mean Average Precision (mAP50) and mean Average Recall (mAR50) at an IoU threshold of 0.5. All results are reported as mean $\pm$ standard deviation over 5 independent runs, with the best performance in each row highlighted in bold. }
  \label{supptable:performance_detection}
\end{table}
\clearpage

\setlength{\tabcolsep}{38pt}
\begin{table}[t]
  \centering
  \begin{adjustbox}{max width=\textwidth}
  \begin{tabular}{lccc}
  \toprule
   & Real & Real+1$\times$Syn. & Real+2$\times$Syn. \\
  \midrule
  \multicolumn{4}{c}{Pneumothorax Segmentation (Internal: SIIM-ACR-PTX)} \\
  \midrule
  Positive Dice & $0.34\pm0.013$ & $0.41\pm0.014$ & $\bm{0.42\pm0.012}$ \\
  Negative Dice & $0.93\pm0.003$ & $0.96\pm0.003$ & $\bm{0.96\pm0.004}$ \\
  \midrule
  \multicolumn{4}{c}{Pneumothorax Segmentation (External: Candi-PTX)} \\
  \midrule
  Positive Dice & $0.28\pm0.015$ & $0.33\pm0.029$ & $\bm{0.36\pm0.020}$ \\
  Negative Dice & $0.98\pm0.002$ & $\bm{0.99\pm0.001}$ & $0.98\pm0.001$ \\
  \midrule
  \multicolumn{4}{c}{Rib Segmentation (Vindr-RibCXR)} \\
  \midrule
  Dice    & $0.80\pm0.002$ & $0.82\pm0.003$ & $\bm{0.83\pm0.001}$ \\
  \bottomrule
  \end{tabular}
  \end{adjustbox}
  \caption{\textbf{Evaluation of synthetic data for training data augmentation in anatomical segmentation.} Segmentation models were trained with real data alone and with synthetic data augmentation (1$\times$ and 2$\times$) for pneumothorax segmentation on SIIM-ACR-PTX (internal) and Candix-PTX (external) datasets, and for rib segmentation on the VinDr-RibCXR dataset. Performance is measured using Dice scores. All results are reported as mean $\pm$ standard deviation over 5 independent runs, with the best performance in each row highlighted in bold. }
  \label{supptable:performance_segmentation}
\end{table}

\clearpage

\begin{table*}[t]
  \centering
  \resizebox{\textwidth}{!}{%
  \begin{tabular}{lccc}
  \toprule
   & Real & Real+1$\times$Syn. & Real+2$\times$Syn. \\
  \midrule
  \multicolumn{4}{c}{RALO} \\
  \midrule
  MAE & $0.65 \pm 0.01$ & $0.62 \pm 0.01$ & $\bm{0.58 \pm 0.01}$ \\
  MSE & $0.71 \pm 0.01$ & $0.68 \pm 0.01$ & $\bm{0.63 \pm 0.01}$ \\
  Pearson'r & $0.675 \pm 0.01$ & $0.69 \pm 0.01$ & $\bm{0.71 \pm 0.01}$ \\
  \bottomrule
  \end{tabular}%
  }
  \caption{\textbf{Evaluation of synthetic data for training data augmentation in severity regression.} Regression models were trained with real data alone and with synthetic data augmentation (1$\times$ and 2$\times$) for opacity severity scoring on the RALO dataset. Performance is measured using Mean Absolute Error (MAE) and Mean Squared Error (MSE). All results are reported as mean $\pm$ standard deviation over 5 independent runs, with the best performance in each row highlighted in bold. }
  \label{supptable:performance_regression}
\end{table*}

\clearpage

\clearpage

\clearpage
%%%%%%%%%%%%%%%%%
%%% References
%%%%%%%%%%%%%%%%%
\begin{nolinenumbers}
\Heading{References}

\vspace{2mm}

\begin{spacing}{0.9}
\bibliographystyle{naturemag}
\bibliography{resources/bibs/main,resources/bibs/dataset}
\end{spacing}
\end{nolinenumbers}
\clearpage

\end{document}